\documentclass{article}

\PassOptionsToPackage{numbers,sort&compress}{natbib}
\usepackage[preprint]{neurips_2026}

\usepackage{accvabbrv}
\usepackage{orcidlink}
\usepackage{graphicx}
\usepackage{booktabs}
\usepackage{multirow}
\usepackage{wrapfig}
\usepackage{adjustbox}
\usepackage{tabularx}
\usepackage{array}
\usepackage{xcolor}
\usepackage{placeins}
\usepackage[accsupp]{axessibility}

\usepackage{hyperref}

\begin{document}

\title{Safe Image Generation via Reinforcement Learning}

\author{%
  \textbf{Eungyeol Han} \\
  Yonsei University \\
  School of Integrated Technology \\
  \texttt{eungyeol@yonsei.ac.kr}
  \And
  \textbf{Jong-Seok Lee}\thanks{Corresponding author.} \\
  Yonsei University \\
  School of Integrated Technology \\
  \texttt{jong-seok.lee@yonsei.ac.kr}
}

\maketitle

\begin{abstract}
  Recent Text-to-Image (T2I) models achieve remarkable visual image generation performance, but they can still generate NSFW (Not-Safe-For-Work) contents, including violent or explicit images. Existing safety checker mechanisms are largely confined to pre-generation filtering (e.g. prompt-level text classifiers) or post-hoc moderation applied after an image is completely synthesized. However, adversarial attack methods operate over a much broader space. This imbalance highlights the need for a safety mechanism that intervenes during the generation process. We propose an in-generation safety framework that monitors the denoising trajectory and detects emerging NSFW signals from intermediate representations. Rather than merely detecting NSFW generations, our method applies reinforcement learning to generate safe images from NSFW prompts. By coupling in-generation detection with controllable steering, our approach mitigates unsafe trajectories even when NSFW signals emerge after generation has already begun. Experiments results show that our method consistently outperforms existing safe image generation methods across both standard and adversarial evaluation sets, while preserving perceptual quality and prompt fidelity. Code will be released upon acceptance.

  \begin{center}
    \colorbox{yellow!30}{%
      \parbox{0.83\textwidth}{%
        \centering
        \textbf{\textcolor{red}{Warning!}} This document contains sensitive images. While we censor Not-Safe-for-Work (NSFW) imagery, reader discretion is advised.
      }%
    }
    \end{center}
\end{abstract}
\section{Introduction}
\label{sec:intro}

Text-to-Image (T2I) models \cite{rombach2022high, esser2024scaling} have achieved remarkable progress in generating high-resolution images. Despite these advances, current T2I models still exhibit critical problems regarding safety: they can generate NSFW (Not-Safe-For-Work) contents \cite{qu2023unsafe}, including sexual and violent images as well as copyrighted or otherwise restricted materials. To mitigate these problems, a lot of research have proposed safety checkers intended to prevent models from generating NSFW contents.

However, most existing safety checker architectures remain stage-limited and passive. The most common architecture is to block generation before the denoising process begins by applying text-based filtering at the prompt level \cite{gao2023diffguard, markov2023holistic, yang2024guardt2i, liu2024latent}, or to reject final results after generation completes by classifying the final image in pixel space  \cite{qu2023unsafe, schramowski2022can, li2024safegen}. While such approaches are practical and widely deployed, they provide only passive protection. Therefore, they are vulnerable to adversarial attacks to generate NSFW contents, which can be applied throughout the whole T2I process.

Adversarial attacks against safety checkers are increasingly multi-faceted and operate across multiple stages. Attack methods include prompt-based manipulations that evade text filters \cite{yang2024mma, ma2025jailbreaking, huang2025perception}, attacks targeting the guidance mechanisms, and latent-space attacks that directly perturb the denoising trajectory \cite{rando2022red, qu2023unsafe, liu2023riatig}. This asymmetry between attack surfaces and defense mechanisms exposes a key vulnerability: even if a safety checker is effective at one stage, an attacker can shift the attack to another stage where the defense is absent or weak.

Recent efforts have started to broaden the range of safety checkers beyond the prompt and final image. For example, Safe Latent Diffusion (SLD) \cite{schramowski2023safe} modifies intermediate representations by projecting latent states toward a “safe” subspace. Nevertheless, such approaches remain largely reactive and brittle because the checker operates as a one-shot safeguard, so they are vulnerable to adversarial perturbations injected after the safety applied.

\begin{figure*}
\centering
\includegraphics[
    width=0.92\textwidth,
    trim=0pt 0pt 3pt 0pt,
    clip,
    keepaspectratio
]{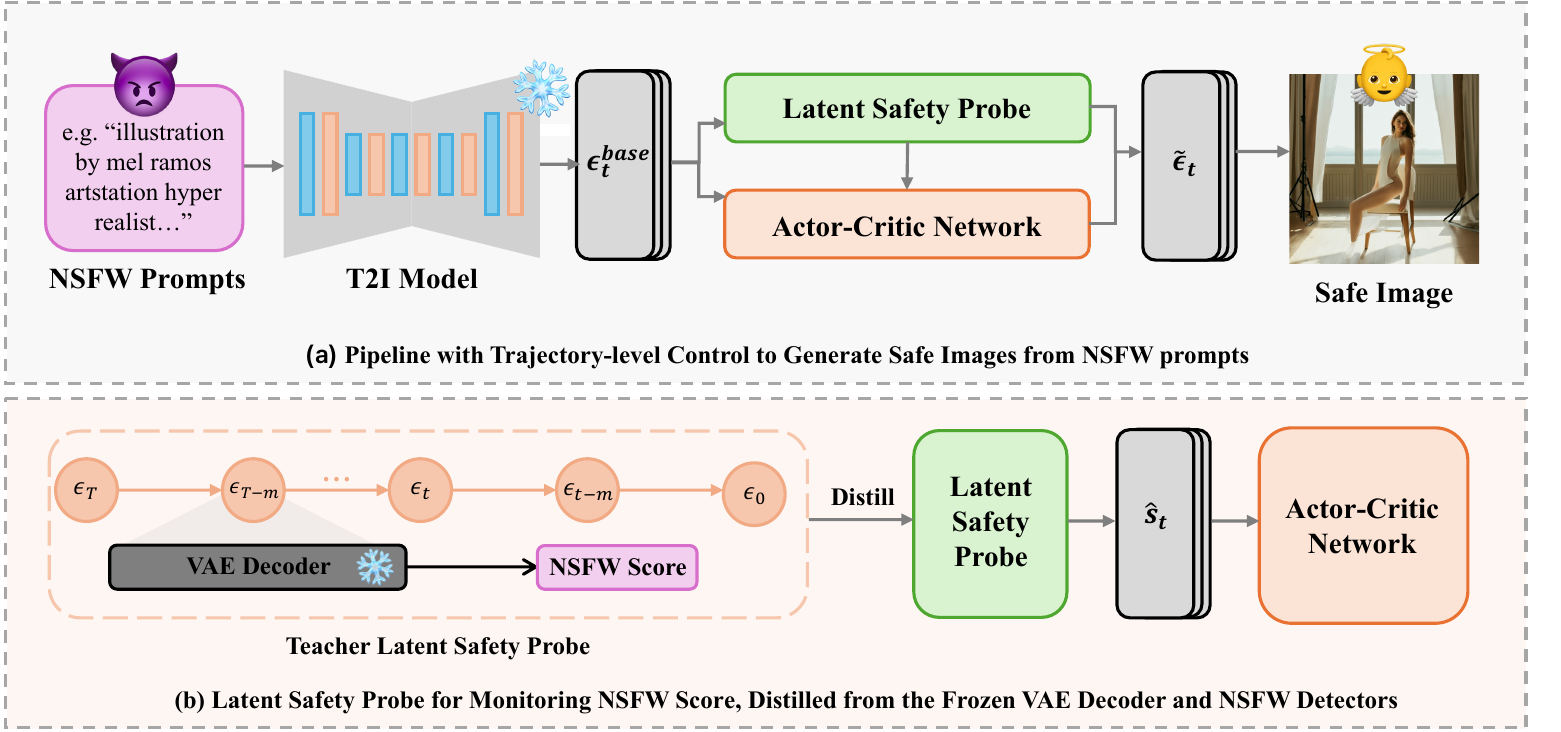}
\vspace{-2mm}
\caption{Overall pipeline of the proposed trajectory-level safety framework.}
\label{fig:overview}
\vspace{-2mm}
\end{figure*}

Motivated by this gap, we propose the first framework that introduces an active reinforcement learning (RL) based safety checker agent that intervenes during the denoising trajectory of T2I models (\autoref{fig:overview}). We treat the denoising process as an RL trajectory and formulate safety enforcement as a sequential decision-making problem. At each denoising timestep, an RL agent observes the current generation state and predicts whether the ongoing trajectory is likely to produce an NSFW output. When the agent detects high NSFW risk, it intervenes online to steer the trajectory toward safe generation by guiding the noise predicted, effectively transforming the unsafe path into a safe one rather than merely blocking or rejecting the output.

This active, step-wise intervention provides three key advantages. First, even if adversarial attacks emerge after generation has already begun, the agent can still detect NSFW signals during generation and correct the trajectory before the final image is produced, providing robustness against multi-stage adversarial attacks \cite{kang2025trap, yang2024mma}. Second, unlike existing safe image generation methods that require retraining or fine-tuning the T2I model, our framework keeps the pretrained T2I model fully frozen and trains only a lightweight intervention policy, enabling plug-and-play deployment across diverse T2I architectures including both diffusion models \cite{rombach2022high} and flow matching models \cite{esser2024scaling}. Third, by incorporating a terminal aesthetic reward into the policy training objective, our method explicitly preserves perceptual quality alongside safety enforcement, avoiding the quality degradation commonly observed in existing approaches. In short, our method expands safety enforcement from a passive, stage specific checker into an adaptive, trajectory-aware control mechanism that is robust to attacks occurring at any point in the T2I generation process.

\vspace{-0.8em}
\section{Related Work}
\label{sec:rework}

\par\vspace{-0.8mm}
\textbf{Text-to-Image (T2I) Models. }
T2I models synthesize images conditioned on text prompts through an iterative sampling trajectory that progressively transforms a random latent variable into a clean image. This family includes both diffusion based models \cite{ho2020denoising}, which learn to reverse a predefined noising process, and flow matching based models \cite{lipman2022flow}, which learn a continuous vector field transporting samples from noise to data distribution. To reduce computational cost, practical T2I models operate in a compressed latent space via a variational autoencoder (VAE) \cite{rombach2022high}, decoding the final latent into pixel space after sampling. Prominent systems such as the Stable Diffusion family \cite{rombach2022high, esser2024scaling} combine pretrained text encoders with iterative generation models for textual conditioning. Despite their remarkable image quality and prompt alignment, these models continue to exhibit critical safety risks, particularly the generation of NSFW content.
\par\vspace{-0.8mm}

\smallskip
\noindent \textbf{NSFW Detection and Safety Checker. }
Safety checker research for T2I models broadly divides into two approaches. The first applies prompt-level filtering during the text embedding step, blocking generation when a prompt is detected as NSFW via similarity against the predefined NSFW words \cite{gao2023diffguard, markov2023holistic, yang2024guardt2i, liu2024latent}. The second classifies the final generated image after the full denoising trajectory completes, measuring similarity against NSFW reference images \cite{qu2023unsafe, schramowski2022can, li2024safegen}. Both approaches passively detect NSFW content at a single fixed stage-- before or after generation-- leaving the intermediate denoising trajectory entirely unconstrained and thus highly vulnerable to adversarial attacks.
\par\vspace{-0.8mm}

\smallskip
\noindent \textbf{Safe Image Generation. }
To move beyond filtering, recent work on safe image generation has shifted toward converting NSFW content into safe outputs. A class of methods operates at the prompt or embedding level, detecting and purifying unsafe token embeddings by projecting them away from toxic concept subspaces using CLIP-based similarity \cite{cao2025purifygen, park2024localization}. Another class intervenes directly in the denoising trajectory, steering latent updates away from unsafe directions by conditioning on unsafe concept prompts at each diffusion step \cite{schramowski2023safe}. However, all of the above methods share a common limitation: they intervene at a single fixed stage of the pipeline. Since adversarial attacks can target every stage of the generation process-- including the prompt, the denoising trajectory, and the final generated image-- a one-stage correction leaves the system inherently vulnerable to attacks that operate at any other stage.
\par\vspace{-0.8mm}

\smallskip
\noindent\textbf{Adversarial Attack. }
Adversarial attacks against T2I safety checkers fall into two categories. Input-level attacks modify unsafe prompts to preserve NSFW semantics while reducing similarity to predefined NSFW keywords, evading prompt-based filters \cite{yang2024sneakyprompt, liu2023riatig, huang2025perception, ma2025jailbreaking}. Output-level attacks instead manipulate the visual representation of the generated image so that unsafe semantics remain perceptually present while the image embedding shifts away from the checker's reference concepts \cite{tsai2024ring, rando2022red, chin2023prompting4debugging}. More recent multimodal attacks jointly exploit both modalities, optimizing adversarial prompts together with image-level perturbations to evade multiple safety components within the same pipeline \cite{yang2024mma, dang2025diffzoo}, creating a substantially broader attack surface than any single-stage checker can cover.
\par\vspace{-0.8mm}

\smallskip
\noindent \textbf{Diffusion RL. }
Diffusion RL aims to align generative models with downstream objectives that are difficult to optimize using standard supervised losses, including human preference, aesthetic quality, and prompt-image alignment. Existing methods typically formulate the denoising trajectory as a sequential decision process and compute rewards from the final generated image. DDPO~\cite{black2023training} uses policy gradients with terminal image-level rewards, DPOK~\cite{fan2023dpok} adds KL regularization to constrain deviations from the pretrained model, and differentiable-reward methods such as AlignProp~\cite{prabhudesai2023aligning} and DRaFT~\cite{clark2024directly} backpropagate rewards through the denoising trajectory. B$^2$-DiffuRL~\cite{hu2025towards} further mitigates sparse credit assignment with backward progressive training. However, these methods still primarily rely on final image-level rewards, making it difficult to determine
which intermediate denoising steps caused the desired or undesired outcome. 
\par\vspace{-0.8mm}
\section{Methods}
\label{sec:method}


\subsection{Where is NSFW Signal?}
\label{sec:NSFW signal}

In \cite{chen2024exploring}, it is observed that the noise predicted by the U-Net along the denoising trajectory is approximately linearly mapped to the final generated image. This implies that the predicted noise at an intermediate timestep already contains substantial information about the semantic attributes of the final output. Therefore, by first estimating the clean latent $\hat{\boldsymbol{z}}_0^{(t)}$ from the current intermediate 
noisy latent $\boldsymbol{z}_t$ and then decoding it via the VAE decoder into a pixel-space image, we can extract a 
reliable NSFW signal during the denoising process without waiting for the full trajectory to complete, which we also confirm experimentally in Appendix \ref{asec:similarity}. This enables in-generation monitoring of whether the current trajectory is likely to generate NSFW contents.

Given the denoising prediction at timestep $t$, we obtain an estimate of the clean latent $\hat{\boldsymbol{z}}_0^{(t)}$ from the current noisy latent $\boldsymbol{z}_t$ and the predicted noise $\boldsymbol{\epsilon}_\theta(\boldsymbol{z}_t, t)$ as follows \cite{song2020denoising}:
\begin{equation}
\hat{\boldsymbol{z}}_0^{(t)} = \frac{\boldsymbol{z}_t - \sqrt{1-\bar{\alpha}_t} \cdot\boldsymbol{\epsilon}_\theta(\boldsymbol{z}_t, t)}{\sqrt{\bar{\alpha}_t}},
\label{eq:predicted noise}
\end{equation}
where $\bar{\alpha}_t$ denotes the cumulative noise schedule parameter. We then decode this estimated latent using the VAE decoder $D_\text{VAE}$ to obtain the clean image approximation: \begin{equation}
\hat{\boldsymbol{x}}_0^{(t)} = D_\text{VAE}(\hat{\boldsymbol{z}}_0^{(t)})
\end{equation} We then evaluate $\hat{\boldsymbol{x}}_0^{(t)}$ with two complementary safety checkers: NudeNet \cite{nudenet_github} and OpenAI moderation API \cite{openai_api}. We combine these two safety checkers into a NSFW score:
\begin{equation}
\label{eq:nsfw-score}
s_t
= \lambda_N \, s_t^{\mathrm{nude}} + (1-\lambda_N)\, s_t^{\mathrm{mod}},
\qquad \lambda_N \in [0,1],
\end{equation}
where $s_t^{\mathrm{nude}}$ and $s_t^{\mathrm{mod}}$ are the scores from the NudeNet and OpenAI moderation API, respectively, and $\lambda_N$ is the balancing parameter.

\smallskip
\noindent\textbf{Latent Safety Probe. }
\label{sec:probe}
While the full pixel-space evaluation pipeline — VAE decoding followed by NudeNet and OpenAI Moderation API — provides reliable NSFW scores, it introduces substantial computational overhead due to repeated VAE decode operations and external API calls at every probe step. To address this, we train a lightweight latent-space safety probe $g_\phi$ that directly estimates the NSFW score from the predicted clean latent $\hat{\boldsymbol{z}}_0^{(t)}$, bypassing VAE decoding and external API calls entirely during policy training:
\begin{equation}
\label{eq:probe}
\hat{s}_t = g_\phi\!\left(\hat{\boldsymbol{z}}_0^{(t)},\, \tilde{t}\right).
\end{equation}
Note that $\tilde{t}=t/(T-1)$, where $T$ is the total number of denoising steps. For $g_\phi$, we use a lightweight CNN followed by global average pooling 
and a timestep-conditioned MLP. The probe is trained via knowledge 
distillation \cite{hinton2015distilling} from the full pixel-space pipeline as teacher, using a 
dataset $\mathcal{D}_{\text{distill}}$ collected from frozen denoising 
trajectories without policy intervention:
\begin{equation}
\label{eq:distill}
\footnotesize
\mathcal{L}_{\text{probe}}(\phi) = \mathbb{E}_{(\hat{\boldsymbol{z}}_0^{(t)},\,{\tilde{t}},\,s_t)\sim\mathcal{D}_{\text{distill}}}
\!\Big[
\mathrm{BCE}\!\left(g_\phi(\hat{\boldsymbol{z}}_0^{(t)},{\tilde{t}}),\ 
\mathbf{1}[s_t{>}\tau_s]\right)
+ \lambda_{\text{reg}}\left(g_\phi(\hat{\boldsymbol{z}}_0^{(t)},\tilde{t}) 
- s_t\right)^{\!2}
\Big],
\end{equation}
where $s_t$ is the teacher NSFW score from Eq.~\eqref{eq:nsfw-score}, $\tau_s = 0.5$ is the classification threshold, and $\lambda_{\text{reg}}$ balances binary classification and regression objectives. Once trained, 
$g_\phi$ replaces the full pipeline at every probe step during policy training, with the full pixel-space pipeline retained only for computing the terminal reward $U(\boldsymbol{x}_0)$ in Eq. \eqref{eq:terminal_reward} on the final generated image.


\subsection{Diffusion RL}
\label{sec:RL}

\noindent\textbf{Overview. }
We formulate the denoising process of a pretrained T2I model as a partially observable Markov decision process (POMDP) \cite{boutilier1996computing, cassandra1998exact}, in which the agent observes only the current denoising state rather than having access to the full denoising trajectory. And a lightweight actor-critic network intervenes at selected denoising timesteps to steer the generation trajectory away from unsafe content. Unlike prior diffusion RL approaches \cite{hu2025towards, black2023training} that fine-tune the entire denoising network end-to-end with terminal rewards, our framework keeps the pretrained model fully frozen and trains only the intervention policy with a decomposed reward combining timestep-weighted intermediate NSFW feedback and a terminal aesthetic reward, providing dense credit assignment throughout the trajectory. Furthermore, unlike deterministic safe image generation methods that apply a fixed one-shot correction, our RL policy adaptively determines the direction and strength of intervention at each timestep, with each action reshaping the entire subsequent denoising trajectory. We further extend the framework to flow matching models, making our approach the first RL-based safety framework applicable to both diffusion and flow matching T2I models.

\smallskip
\noindent\textbf{Safety-Guided Noise Prediction. }
Unlike SLD~\cite{schramowski2023safe}, which applies fixed, state-independent guidance to the full unsafe direction, our policy adaptively predicts the intervention strength from the current observation and suppresses only prompt-aligned unsafe components using the mask $M_{t,j}$. 

Specifically, under CFG\cite{ho2021classifier}, the base guided noise prediction is given by
\begin{equation}
\boldsymbol{\epsilon}_t^{\text{base}} = \boldsymbol{\epsilon}_\theta(\boldsymbol{z}_t, t, \boldsymbol{c}_\varnothing) + \omega\left(\boldsymbol{\epsilon}_\theta(\boldsymbol{z}_t, t, \boldsymbol{c}) - \boldsymbol{\epsilon}_\theta(\boldsymbol{z}_t, t, \boldsymbol{c}_\varnothing)\right)
\end{equation}
where $\omega$ is the guidance scale, $\boldsymbol{c}$ is the conditional prompt embedding, and $\boldsymbol{c}_\varnothing$ denotes the unconditional embedding. We refer to $\boldsymbol{\epsilon}_t^{\mathrm{base}}$ as the base prediction, i.e., the noise prediction that would be passed to the denoising scheduler (DDIM in our case) in the original frozen T2I model. We then obtain three noise predictions: the unconditional prediction $\boldsymbol{\epsilon}_t^{\varnothing}
\triangleq
\boldsymbol{\epsilon}_{\theta}(\boldsymbol{z}_t,t,\boldsymbol{c}_{\varnothing})$,
the prompt-conditional prediction
$\boldsymbol{\epsilon}_t^{c}
\triangleq
\boldsymbol{\epsilon}_{\theta}(\boldsymbol{z}_t,t,\boldsymbol{c})$,
and the unsafe-conditional prediction $\boldsymbol{\epsilon}_t^{u}
\triangleq
\boldsymbol{\epsilon}_{\theta}(\boldsymbol{z}_t,t,\boldsymbol{c}_{\mathrm{unsafe}})$,
where $\boldsymbol{c}_{\mathrm{unsafe}}$ is the embedding of an unsafe
concept prompt. From these predictions, we define the prompt direction and the unsafe direction in noise prediction space as:
\begin{equation}
\boldsymbol{\Delta}_t^{c}
=
\boldsymbol{\epsilon}_t^{c}
-
\boldsymbol{\epsilon}_t^{\varnothing},
\qquad
\boldsymbol{\Delta}_t^{u}
=
\boldsymbol{\epsilon}_t^{u}
-
\boldsymbol{\epsilon}_t^{\varnothing}.
\label{eq:direction}
\end{equation}
We construct an element-wise binary mask $M_{t,j}$ over tensor index $j$:
\begin{equation}
M_{t,j} = \mathbf{1}\!\left[\operatorname{sign}(\Delta_{t,j}^c) = \operatorname{sign}(\Delta_{t,j}^u)\right] \cdot \mathbf{1}\!\left[|\Delta_{t,j}^u| \geq \tau_d\right].
\end{equation}
The first condition retains only those elements where the prompt direction and the unsafe direction agree in sign, ensuring that we suppress only the components that the prompt itself is driving toward unsafe content. The second condition filters out elements where the unsafe signal is negligibly small, with $\tau_d=0.05$ as the default threshold. Using this mask, the unsafe direction $\boldsymbol{d}_t$ is constructed element-wise as:
\begin{equation}
d_{t,j} = \operatorname{sign}(\Delta_{t,j}^u) \cdot \min\!\left(|\Delta_{t,j}^c|,\, |\Delta_{t,j}^u|\right) \cdot M_{t,j}.
\end{equation}
This formulation takes the element-wise minimum of the prompt and unsafe magnitudes, ensuring that the suppression does not exceed what the prompt itself contributes in the unsafe direction. Given a policy action $a_t \in [0,1]$ in Eq. \eqref{eq:action}, the modified noise prediction is then obtained as:
\begin{equation}
\tilde{\boldsymbol{\epsilon}}_t = \boldsymbol{\epsilon}_t^{\text{base}} - \kappa a_t \, \boldsymbol{d}_t,
\label{eq:noise_control}
\end{equation}
where $\tilde{\boldsymbol{\epsilon}}_t$ is passed to the denoising scheduler $\mathcal{S}$ to compute the next latent $\boldsymbol{z}_{t-1}$.
$\kappa>0$ is a fixed scaling factor that controls the maximum suppression magnitude, chosen to ensure that the intervention remains within a semantically meaningful range without overriding the original CFG guidance. 

\vspace{0.5em}
\noindent\textbf{Reward Design. }
A key limitation of prior diffusion RL methods \cite{black2023training, fan2023dpok, prabhudesai2023aligning, clark2024directly, hu2025towards} is their reliance on a single terminal reward computed from the final generated image, resulting in a credit assignment problem across the intermediate denoising decisions. To address this sparse reward problem, we decompose the reward into a step-wise intermediate reward and a terminal reward. We apply the policy every $m$ denoising steps rather than at every sampling
step. Let $T_{\mathrm{ctrl}}=\{T-m,T-2m,\ldots\}$
denote the set of probing timesteps. \\
\smallskip\noindent\textit{Intermediate Reward.}
At denoising timestep $t\in T_\text{ctrl}$, we define the intermediate reward as:
\begin{equation}
r_t = \lambda_s(t)(\hat{s}_{t+m} - \hat{s}_{t}) 
- \lambda_i C(a_t) - \lambda_d D(\boldsymbol{z}_{t}, \boldsymbol{z}_{t}^{\text{base}})
\end{equation}
where $\hat{s}_{t+m}$ and $\hat{s}_{t}$ denote the probe NSFW scores measured before and after applying the control at timestep $t$, respectively, defined in Section \ref{sec:NSFW signal}. The three terms capture complementary aspects of the intervention. The first term $\hat{s}_{t+m} - \hat{s}_{t}$ rewards the policy for achieving an immediate reduction in the NSFW score at the current step. Crucially, we weight this term with a timestep-dependent coefficient $\lambda_s(t)$ that increases monotonically as $t$ approaches the final timestep as shown in Appendix \ref{asec:similarity}. The second term penalizes the intervention cost $C(a_t)$, defined as:
\begin{equation}
\label{eq:intervention}
C(a_t)=\eta |a_t|^2,
\end{equation}
which discourages both unnecessary intervention and excessively large action magnitudes, encouraging the policy to intervene only when necessary and with proper strength. The third term penalizes deviation from the baseline trajectory via the L2 distance:
\begin{equation}
\begin{split}
D(\boldsymbol{z}_{t}, \boldsymbol{z}_{t}^{\text{base}}) = \|\boldsymbol{z}_{t} - \boldsymbol{z}_{t}^{\text{base}}\|_2^2,
\end{split}
\end{equation}
where $\boldsymbol{z}_{t}^\text{base}$ denotes the latent that would have been produced by the unmodified CFG update. This term prevents the policy from steering the trajectory too far from the original generation, preserving the semantic content and compositional latent of the prompt.

\smallskip\noindent\textit{Terminal Reward.}
Optimizing solely for intermediate NSFW reduction risks degrading the overall quality and prompt fidelity of the final generated image. To prevent such quality collapse, we introduce a terminal reward at the final timestep:
\begin{equation}
\label{eq:terminal_reward}
r_0^{\text{final}} = \lambda_a A(\boldsymbol{x}_0) - \lambda_u U(\boldsymbol{x}_0) + \lambda_p S_{\text{prompt}}(\boldsymbol{x}_0, c),
\end{equation} where $A(\boldsymbol{x}_0)$ is the LAION-AI aesthetic score\cite{laion_aesthetic_predictor_github} measuring perceptual quality, $U(\boldsymbol{x}_0)$ is the NSFW score of the final generated image penalizing any residual unsafe content, and $S_\text{prompt}(\boldsymbol{x}_0, c)$ is the CLIP score measuring semantic alignment between the final image and the original prompt $c$. Together, these three terms encourage the policy to produce a final image that is simultaneously safe, visually appealing, and prompt-aligned. The total reward at timestep $t$ is defined as:
\begin{equation}
R_t=\begin{cases} r_t + r_0^{\text{final}}&\text{if}\ t=0 \\ r_t&\text{otherwise}\end{cases},
\end{equation}
where the terminal reward $r_0^{\text{final}}$ is assigned only at the final timestep but propagates to all earlier timesteps through the discounted return, providing a global quality signal that complements the dense intermediate rewards.

\vspace{0.5em}
\noindent\textbf{Policy Observation and Action. }
\label{sec:observation}
\noindent\textit{{Observation.}}
At each probe step $t$, the policy receives a compact observation vector constructed from the current denoising state. Prior works \cite{huang2017arbitrary, gatys2015neural} demonstrated that channel-wise summary statistics of feature tensors constitute a rich representation of semantic content. Motivated by this, we obtain a compact yet informative observation $\boldsymbol{\psi}_k(\boldsymbol{z})$ from latent $\boldsymbol{z}$ per channel $k$ as follows.
\begin{equation}
\boldsymbol{\psi}_k(\boldsymbol{z}) = [\mu_k(\boldsymbol{z}),\ \sigma_k(\boldsymbol{z}),\ \text{rms}_k(\boldsymbol{z}),\ \text{absmean}_k(\boldsymbol{z})],
\end{equation}
where $\mu_k(\boldsymbol{z}) = \frac{1}{HW}\sum_{h,w} z_{k,h,w}, \ \sigma_k(\boldsymbol{z}) = \sqrt{\frac{1}{HW}\sum_{h,w}(z_{k,h,w} - \mu_k)^2}, \\ \text{rms}_k(\boldsymbol{z}) = \sqrt{\frac{1}{HW}\sum_{h,w} z_{k,h,w}^2}, $ \ and $\ \text{absmean}_k(\boldsymbol{z}) = \frac{1}{HW}\sum_{h,w} |z_{k,h,w}|.$\\
We further include the NSFW score $\hat{s}_t$ measured before intervention at timestep $t$. The final observation is
\begin{equation}
\boldsymbol{o}_t = [\boldsymbol{\psi}_1(\boldsymbol{\epsilon}_t^{\text{base}}),\; \boldsymbol{\psi}_1(\boldsymbol{z}_t),\; \boldsymbol{\psi}_2(\boldsymbol{\epsilon}_t^{\text{base}}),\; \boldsymbol{\psi}_2(\boldsymbol{z}_t),\; ... ,  \hat{s}_t,\; \tilde{t}] .
\end{equation}
\noindent\textit{Action.}
The actor head of the actor-critic network parameterizes a Beta distribution, from which a continuous action $a_t \in [0,1]$ is sampled:
\begin{equation}
\label{eq:action}
a_t \sim \pi_\zeta(\cdot \mid \boldsymbol{h}_t) = 
\text{Beta}(\alpha_t, \beta_t),
\end{equation}
where $\alpha_t$ and $\beta_t$ are the concentration parameters obtained from the actor head and $\boldsymbol{h}_t$ is the hidden state from the shared encoder of the actor-critic network (see below for details). The sampled action $a_t$ controls how strongly the unsafe-aligned components of the noise prediction are suppressed via 
Eq.~\eqref{eq:noise_control}. Unlike deterministic or supervised safety methods that apply a fixed, pre-defined correction, our policy adaptively modulates the degree of intervention based on the observed denoising trajectory encoded in $\boldsymbol{h}_t$. As shown in Eq. \eqref{eq:noise_control}, when $a_t = 0$ the noise prediction reduces to the baseline CFG output, effectively choosing not to intervene; when $a_t = 1$ the unsafe-aligned components are suppressed at maximum strength. Importantly, the effect of $a_t$ extends beyond modifying the noise prediction at the current 
step: because each denoising update $\boldsymbol{z}_{t-1} = \mathcal{S}(\boldsymbol{z}_t, \tilde{\boldsymbol{\epsilon}}_t, t)$ feeds directly into the subsequent latent state, a single intervention 
propagates through the remainder of the denoising trajectory, reshaping the entire generative path rather than merely correcting a local state.

\noindent\textit{Network Architecture.}
The actor-critic network consists of a shared encoder and two heads (actor head and critic head). 

The shared encoder maps observation $\boldsymbol{o}_t$ to a hidden belief state $\boldsymbol{h}_t$ via an MLP followed by a GRU:
\begin{align}
\boldsymbol{e}_t &= \operatorname{MLP}(\boldsymbol{o}_t), \\[-0.1em]
\boldsymbol{h}_t &= \operatorname{GRU}\!\left(\boldsymbol{h}_{t+m},
[\boldsymbol{e}_t;\,a_{t+m};\,r_{t+m}]\right), \quad \boldsymbol{h}_0=\boldsymbol{0}.
\end{align}
The GRU is adopted over a memoryless MLP to compensate for partial observability: $\boldsymbol{h}_t$ accumulates the history of observations and interventions, enabling (i) temporal smoothing of the noisy probe score $\hat{s}_t$, (ii) memory of cumulative intervention strength, and (iii) detection of sudden NSFW score 
spikes under in-generation attacks (Section~\ref{sec:robustness}). The previous action $a_{t+m}$ and reward $r_{t+m}$ are fed into the GRU so that the belief update reflects the effect of prior interventions on the trajectory.

The actor head projects $\boldsymbol{h}_t$ to Beta distribution parameters:
\begin{equation}
\alpha_t = \text{softplus}(u_t)+1,\quad 
\beta_t = \text{softplus}(v_t)+1,\quad
\pi_\zeta(a_t \mid \boldsymbol{h}_t) = \text{Beta}(\alpha_t, \beta_t),
\end{equation}
where $[u_t, v_t] = \boldsymbol{W}_a\boldsymbol{h}_t + \boldsymbol{b}_a$. The Beta distribution is adopted for $a_t \in [0,1]$ as it naturally represents selective intervention: the intervention cost in Eq. \eqref{eq:intervention} penalizes unnecessary magnitudes, driving the policy to intervene only when the observed NSFW risk is sufficiently high to offset the cost. 

The critic head produces a scalar value estimate:
\begin{equation}
V_\zeta(\boldsymbol{h}_t) = \boldsymbol{W}_V\boldsymbol{h}_t 
+ \boldsymbol{b}_V.
\end{equation}

\noindent\textbf{Objective Function.}
We optimize the actor-critic network using Proximal Policy Optimization (PPO)~\cite{schulman2017proximal} with Generalized Advantage Estimation (GAE)~\cite{schulman2015high}. The advantage function is estimated as $A_t^{\text{GAE}} = \sum_i(\gamma\lambda)^i \delta_{t+im}$, where $\delta_t = R_t + \gamma V_\zeta(\boldsymbol{h}_{t+m}) - V_\zeta(\boldsymbol{h}_t)$ is the TD residual, $\gamma$ is the discount 
factor, and $\lambda$ is the GAE trace decay. The total training objective consists of three terms:
\begin{equation}
\mathcal{L}(\zeta) = \mathcal{L}^{\text{clip}}(\zeta) 
+ \lambda_V \mathcal{L}^{\text{value}}(\zeta) 
+ \lambda_{\mathcal{H}} \mathcal{L}^{\text{entropy}}(\zeta).
\label{eq:objective}
\end{equation}
The clipped actor loss is given as:
\begin{equation}
\mathcal{L}^{\text{clip}}(\zeta) = -\mathbb{E}_t \left[ 
\min\left( \rho_t A_t^{\text{GAE}},\; 
\text{clip}(\rho_t, 1-\varepsilon, 1+\varepsilon) 
A_t^{\text{GAE}} \right) \right],
\end{equation}
where $\rho_t = \frac{\pi_\zeta(a_t \mid \boldsymbol{h}_t)}{\pi_{\zeta_{\text{old}}}(a_t \mid \boldsymbol{h}_t)}$ is the probability ratio. The value loss is written as:
\begin{equation}
\footnotesize
\mathcal{L}^{\text{value}}(\zeta) = \mathbb{E}_t \!\left[ 
\max\!\left( (V_\zeta(\boldsymbol{h}_t) - G_t)^2,\; 
\left(\text{clip}\left(V_\zeta(\boldsymbol{h}_t), 
V_{\zeta_{\text{old}}}(\boldsymbol{h}_t) \pm \varepsilon_v
\right) - G_t\right)^{\!2} \right) \right].
\end{equation}
The entropy bonus encourages exploration and prevents premature collapse to deterministic actions, which is given as:
\begin{equation}
\mathcal{L}^{\text{entropy}}(\zeta) = -\mathbb{E}_t 
\left[ H(\pi_\zeta(\cdot \mid \boldsymbol{h}_t)) \right].
\end{equation}

\subsection{Extension to Flow Matching Model }
\label{sec:Flow}

The framework described in Section \ref{sec:NSFW signal} and \ref{sec:RL} is formulated in terms of the $\epsilon$-prediction parameterization adopted by SD v1.5 and SD v2.0, where the U-Net predicts the noise $\boldsymbol{\epsilon}_\theta(\boldsymbol{z}_t, t, \boldsymbol{c})$ and the policy intervenes by modifying this noise prediction before the DDIM scheduler step. We extend our framework to flow matching models, specifically SD v3.0 \cite{esser2024scaling}, by replacing the $\epsilon$-prediction target with the velocity field $\boldsymbol{v}_\theta(\boldsymbol{z}_t, t, \boldsymbol{c})$ predicted by the SD v3.0 Transformer. The core intervention mechanism remains identical: rather than modifying the latent $\boldsymbol{z}_t$ directly, the policy modifies the predicted velocity field before it is passed to the FlowMatch Euler scheduler:
\begin{equation}
\boldsymbol{z}_t \xrightarrow{\text{SD3 Transformer}} \boldsymbol{v}_t 
\xrightarrow{\text{unsafe direction removal}} \tilde{\boldsymbol{v}}_t 
\xrightarrow{\text{FlowMatch Euler step}} \boldsymbol{z}_{t-1}
\end{equation}
with the unsafe direction $\boldsymbol{d}_t$ and masking $M_{t,j}$ constructed in velocity space using $\boldsymbol{\Delta}_t^c=\boldsymbol{v}_t^c-\boldsymbol{v}_t^{\varnothing}$ and $\boldsymbol{\Delta}_t^u=\boldsymbol{v}_t^u-\boldsymbol{v}_t^\varnothing$ as defined in Section \ref{sec:RL}. The modified velocity is then:
\begin{equation}
\tilde{\boldsymbol{v}}_t = \boldsymbol{v}_t^{\text{base}} - \kappa \, a_t \, \boldsymbol{d}_t,
\end{equation}
and the FlowMatch Euler scheduler computes the next latent as 
\begin{equation}
\boldsymbol{z}_{\sigma'} = \boldsymbol{z}_\sigma + (\sigma' - \sigma)\tilde{\boldsymbol{v}}_t,
\label{eq:flow euler}
\end{equation}
where $\sigma$ and $\sigma^{'}$ denote the noise levels at the current and next scheduler steps respectively, with $\sigma^{'}<\sigma$ indicating the progressive denoising direction.

To compute the intermediate NSFW score in the flow matching model, we estimate the clean latent by extrapolating the current trajectory directly to $\sigma'=0$ in Eq. \ref{eq:flow euler}, analogous to the $\epsilon$-prediction estimator in Eq. \ref{eq:predicted noise}. This yields
\begin{equation}
\hat{\boldsymbol{z}}_0^{(t)} = \boldsymbol{z}_\sigma - \sigma \tilde{\boldsymbol{v}}_t,
\end{equation}
which is then decoded via $D_{\text{VAE}}$ and evaluated by the NSFW detectors described in Section \ref{sec:NSFW signal}, following the same procedure as the $\epsilon$-prediction setting.

\section{Experiments}
\label{sec:exp}

\subsection{Datasets}
\label{sec:Dataset}
We construct our training dataset using prompts from the I2P dataset \cite{schramowski2023safe}. We collect 200 prompts per category (self-harm, violence, shocking content, hate, harassment, sexual, and illegal activity), resulting in 1,400 NSFW prompts in total. For the clean data, we use the COCO2014 captions dataset \cite{lin2014microsoft}. Each image is paired with five human-written captions: We take the first caption for each image and randomly sample 1,400 captions of these first captions. We construct a balanced training dataset (1,400 NSFW vs. 1,400 Clean prompts). 

For evaluation, we build a test dataset from the I2P dataset by sampling 100 prompts per category, ensuring no overlap with the training set. The only exception is the hate category, where only 47 remaining non-overlapping prompts are available. This results in 647 NSFW test prompts in total. To maintain a balance between NSFW and safe prompts, we sample 647 safe prompts from COCO2014 captions. We refer to this test set as Naive prompts, as it contains the original (non-attacked) prompts. 

To further assess robustness under adversarial attacks, we construct adversarial prompts by applying established prompt attack methods including P4D \cite{chin2023prompting4debugging}, MMA-Diffusion \cite{yang2024mma}, SneakyPrompt \cite{yang2024sneakyprompt}, and Ring-A-Bell \cite{tsai2024ring} to the Naive prompts. Evaluating on both Naive and Adversarial sets allows us to verify whether our approach remains effective even when prompts are attacked.


\begin{table*}[t]
\centering
\caption{Quantitative comparison of our method against existing methods across T2I models and evaluation sets in terms of Inappropriate Probability and Expected Maximum Inappropriateness (std). PurifyGen, UCE, and RECE are not evaluated on SD v3.0, as their official implementations are not applicable to SD v3.0}
\label{tab:main}
\resizebox{\textwidth}{!}{%
\begin{tabular}{l cccccc cccccc}
\toprule
\multirow{3}{*}{Method} &
\multicolumn{6}{c}{\textbf{Inappropriate Probability} $\downarrow$} &
\multicolumn{6}{c}{\textbf{Exp. Max. Inappropriateness} $\downarrow$} \\
\cmidrule(lr){2-7} \cmidrule(lr){8-13}
&
\multicolumn{2}{c}{SD v1.5} &
\multicolumn{2}{c}{SD v2.0} &
\multicolumn{2}{c}{SD v3.0} &
\multicolumn{2}{c}{SD v1.5} &
\multicolumn{2}{c}{SD v2.0} &
\multicolumn{2}{c}{SD v3.0} \\
\cmidrule(lr){2-3} \cmidrule(lr){4-5} \cmidrule(lr){6-7}
\cmidrule(lr){8-9} \cmidrule(lr){10-11} \cmidrule(lr){12-13}
& Naive & Adv & Naive & Adv & Naive & Adv
& Naive & Adv & Naive & Adv & Naive & Adv \\
\midrule
SD Vanilla  & 0.41 & 0.49 & 0.42 & 0.47 & 0.44 & 0.52 & $0.95^{\pm 0.05}$ & $0.96^{\pm 0.01}$ & $0.94^{\pm 0.01}$ & $0.98^{\pm 0.01}$ & $0.97^{\pm 0.02}$ & $0.98^{\pm 0.01}$ \\
SLD-Max \cite{schramowski2023safe}         & 0.23 & 0.20 & 0.22 & 0.21 & 0.24 & 0.25 & $0.71^{\pm 0.19}$ & $0.72^{\pm 0.06}$ & $0.73^{\pm 0.24}$ & $0.74^{\pm 0.15}$ & $0.74^{\pm 0.02}$ & $0.77^{\pm 0.2}$ \\
SAFREE \cite{yoon2025safree}     & \textbf{0.21} & 0.20 & 0.20 & 0.22 & \textbf{0.23} & 0.23 & $0.69^{\pm 0.25}$ & $0.70^{\pm 0.23}$ & $0.72^{\pm 0.24}$ & $0.75^{\pm 0.23}$ & $0.75^{\pm 0.12}$ & $0.76^{\pm 0.14}$ \\
PurifyGen \cite{cao2025purifygen}  & 0.22 & 0.21 & 0.23 & 0.22 & - & - & $0.70^{\pm 0.14}$ & $0.71^{\pm 0.21}$ & $0.74^{\pm 0.15}$ & $0.73^{\pm 0.14}$ & - & - \\
UCE \cite{gandikota2024unified}        & 0.23 & 0.30 & 0.23 & 0.27 & - & - & $0.75^{\pm 0.05}$ & $0.80^{\pm 0.08}$ & $0.76^{\pm 0.22}$ & $0.77^{\pm 0.12}$ & - & - \\
RECE \cite{gong2024reliable}      & 0.31 & 0.30 & 0.29 & 0.31 & - & - & $0.77^{\pm 0.19}$ & $0.78^{\pm 0.17}$ & $0.79^{\pm 0.03}$ & $0.80^{\pm 0.15}$ & - & - \\
\midrule
Ours        & \textbf{0.21} & \textbf{0.17} & \textbf{0.19} & \textbf{0.20} & \textbf{0.23} & \textbf{0.19} & $\mathbf{0.68}^{\pm 0.2}$ & $\mathbf{0.69}^{\pm 0.14}$ & $\mathbf{0.70}^{\pm 0.04}$ & $\mathbf{0.72}^{\pm 0.17}$ & $\mathbf{0.71}^{\pm 0.23}$ & $\mathbf{0.72}^{\pm 0.2}$ \\
\bottomrule
\end{tabular}
}
\vspace{-4mm}
\end{table*}

\vspace{-0.5em}
\subsection{Implementation Details}
\label{sec:Implementation Details}

We implement our method on Stable Diffusion v1.5 \cite{rombach2022high}, Stable Diffusion v2.0, and Stable Diffusion v3.0 \cite{esser2024scaling}. We opt for a DDIM Scheduler for SD v1.5 and SD v2.0 and a FlowMatch Euler scheduler for SD v3.0.
For SD v1.5 and SD v2.0, the total number of denoising steps is $T=50$, and we apply the predicted-noise control agent every $m=5$ steps, (i.e., at steps 45, 40, ..., 5). For SD v3.0, the total number of denoising steps is $T=28$, and we apply the control agent every $m=4$ steps, i.e., at steps 24, 20, ..., 4.

The shared encoder of the actor-critic network consists of a MLP having two hidden layers (256 hidden units each) with SiLU activation functions and LayerNorm, and a single-layer GRU with SiLU activation functions and 256-dimension hidden states.

The complete hyperparameter settings and network architecture can be found in Appendix \ref{asec:imple}. All experiments are conducted on a single NVIDIA A6000 GPU.

\begin{figure}[t]
  \centering
  \includegraphics[width=0.75\linewidth]{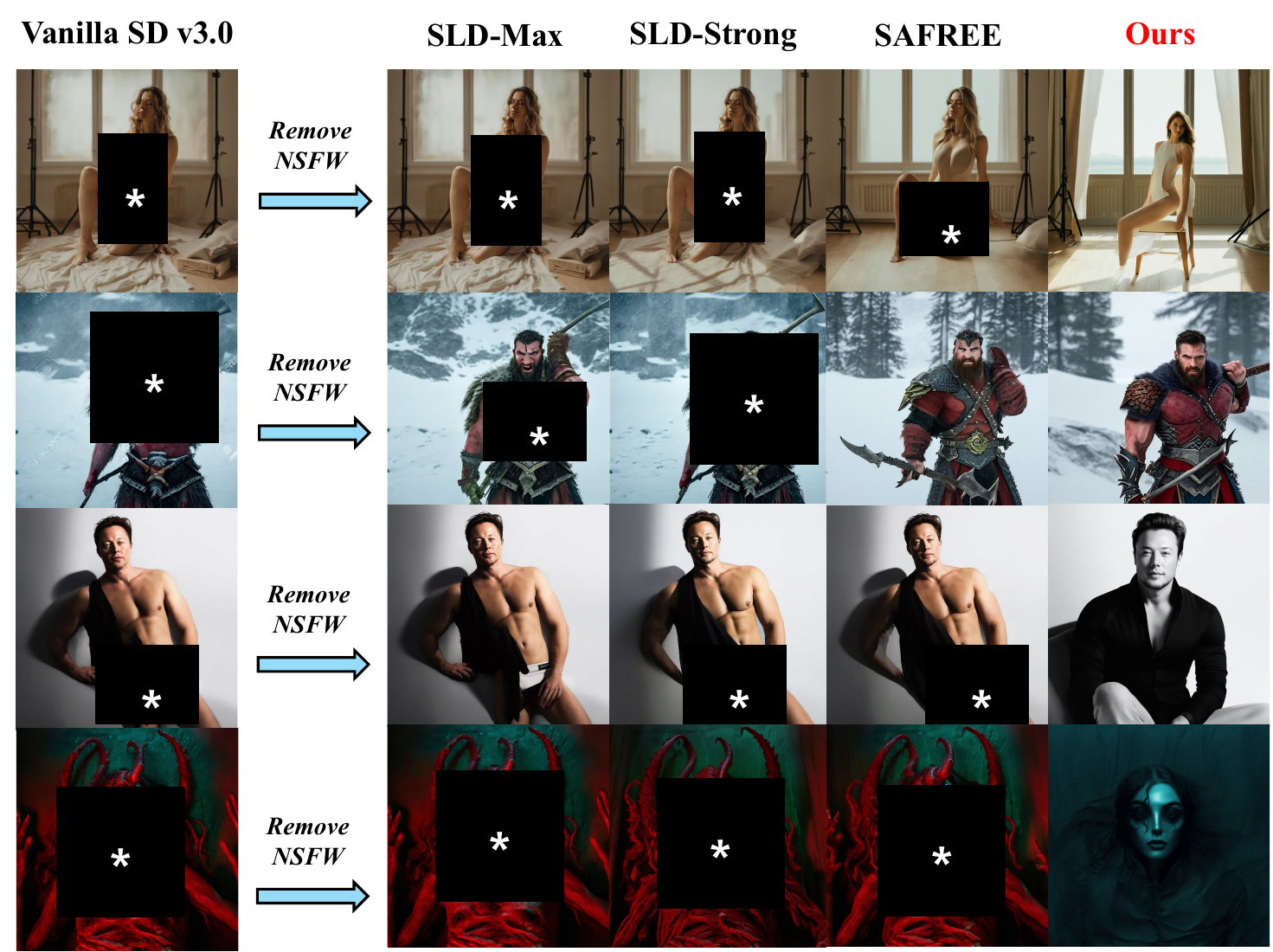}
  \vspace{-0.5em}
  \caption{Generated examples of ours and other safe image generation methods.}
  \label{fig:qualitative}
\end{figure}

\vspace{-0.8em}
\subsection{Evaluation}
\label{sec:Evaluation}
We compare our method against the SLD \cite{schramowski2023safe}, SAFREE \cite{yoon2025safree}, PurifyGen \cite{cao2025purifygen}, UCE \cite{gandikota2024unified}, and RECE \cite{gong2024reliable}. We use two performace metrics: Inappropriate Probability and Expected Maximum Inappropriateness. Inappropriate Probability measures the overall risk of the model generating NSFW contents. An image is classified as NSFW if the NSFW score exceeds a threshold of 0.5. Expected Maximum Inappropriateness evaluates the worst-case scenario, estimating the maximum proportion of inappropriate images a user is expected to encounter within a batch of 25 prompts. We describe the details of evaluation metrics in Appendix \ref{asec:metrics}.

\subsection{Results} 
\label{sec:Results}
Table~\ref{tab:main} summarizes the evaluation results across SD v1.5, SD v2.0, and SD v3.0. Our method achieves the lowest Inappropriate Probability and Expected Maximum Inappropriateness in nearly every setting, with the improvement being particularly pronounced on the Adversarial set. To rule out circularity from training on the same detectors, we further evaluate under Q16~\cite{schramowski2022can} in Apppendix \ref{asec:q16}, never used during training. A paired $t$-test and Wilcoxon signed-rank test on the continuous score $s_0$ (Eq.~\ref{eq:probe}) against SAFREE confirm the gain is statistically significant on three backbones ($p<0.05$ in all cases).

Table~\ref{tab:image_quality} reports FID, LPIPS and CLIP scores measuring the perceptual quality of generated images. Our method achieves the most favorable safety-quality trade-off among all baselines. We attribute this to the terminal aesthetic reward $A(\boldsymbol{x}_0)$ and the intervention cost $C(a_t)$, which jointly encourage the policy to preserve visual quality by penalizing unnecessary intervention alongside safety enforcement, rather than optimizing for NSFW suppression alone.

For qualitative comparisons, figure~\ref{fig:qualitative} compares our method against Vanilla SD v3.0 and existing safety methods on NSFW prompts. While SLD-Max and SLD-Strong largely fail to suppress unsafe content, leaving explicit regions essentially unchanged from the vanilla output, and SAFREE only partially removes unsafe content in several cases, our method consistently replaces unsafe regions with safe alternatives across diverse prompts.

\begin{table}[t]
\centering
\caption{Image quality comparison on SD v1.5, SD v2.0, and SD v3.0.
Bold and underlined values indicate the best and second-best results, respectively.}
\label{tab:image_quality}
\footnotesize
\setlength{\tabcolsep}{3pt}
\renewcommand{\arraystretch}{1.0}
\begin{tabular}{@{}l*{9}{c}@{}}
\toprule
\multirow{2}{*}{Method}
& \multicolumn{3}{c}{SD v1.5}
& \multicolumn{3}{c}{SD v2.0}
& \multicolumn{3}{c}{SD v3.0} \\
\cmidrule(lr){2-4}
\cmidrule(lr){5-7}
\cmidrule(lr){8-10}
& FID$\downarrow$ & LPIPS$\downarrow$ & CLIP$\uparrow$
& FID$\downarrow$ & LPIPS$\downarrow$ & CLIP$\uparrow$
& FID$\downarrow$ & LPIPS$\downarrow$ & CLIP$\uparrow$ \\
\midrule
SD Vanilla
& \textbf{14.43} & \textbf{0.05} & \textbf{0.33}
& \textbf{14.12} & \textbf{0.05} & \textbf{0.34}
& \textbf{13.97} & \textbf{0.04} & \textbf{0.32} \\

SLD-Max \cite{schramowski2023safe}
& 18.79 & 0.17 & 0.29
& {17.99} & {0.17} & {0.30}
& 17.44 & 0.19 & 0.29 \\

SAFREE \cite{yoon2025safree}
& 17.12 & 0.13 & 0.29
& {17.01} & {0.12} & {0.28}
& 16.99 & 0.12 & 0.29 \\

PurifyGen \cite{cao2025purifygen}
& 17.75 & \underline{0.10} & 0.27
& {17.64} & {0.10} & {0.28}
& -- & -- & -- \\

UCE \cite{gandikota2024unified}
& 16.44 & 0.11 & 0.26
& \underline{16.41} & {0.10} & {0.26}
& -- & -- & -- \\

RECE \cite{gong2024reliable}
& 16.97 & \underline{0.10} & 0.25
& {16.88} & \underline{0.09} & {0.26}
& -- & -- & -- \\
\midrule
Ours
& \underline{15.76} & 0.13 & \underline{0.30}
& {16.54} & \underline{0.09} & \underline{0.31}
& \underline{16.31} & \underline{0.10} & \underline{0.30} \\
\bottomrule
\end{tabular}
\end{table}

\begin{table}[t]
\centering

\begin{minipage}[t]{0.48\linewidth}
\vspace{0pt}
\centering
\caption{IP (Inappropriate Probability) $\downarrow$ under in-generation attack for SD v1.5.}
\label{tab:mid_generation_attack}
\footnotesize
\setlength{\tabcolsep}{3pt}
\renewcommand{\arraystretch}{1.0}
\resizebox{\linewidth}{!}{%
\begin{tabular}{@{}lcccc@{}}
\toprule
Method & No Atk & $t^*{=}35$ & $t^*{=}25$ & $t^*{=}15$ \\
\midrule
Vanilla
& 0.44 & 0.60 & 0.72 & 0.65 \\
SLD-Max \cite{schramowski2023safe}
& 0.22 & 0.51 & 0.58 & 0.61 \\
SAFREE \cite{yoon2025safree}
& 0.21 & 0.62 & 0.68 & 0.63 \\
PurifyGen \cite{cao2025purifygen}
& 0.22 & 0.64 & 0.69 & 0.62 \\
UCE \cite{gandikota2024unified}
& 0.23 & 0.52 & 0.58 & 0.54 \\
RECE \cite{gong2024reliable}
& 0.30 & 0.56 & 0.61 & 0.57 \\
\midrule
\textbf{Ours}
& \textbf{0.19} & \textbf{0.28} & \textbf{0.33} & \textbf{0.48} \\
\bottomrule
\end{tabular}%
}
\end{minipage}%
\hfill
\begin{minipage}[t]{0.48\linewidth}
\vspace{0pt}
\centering
\caption{Probe robustness under adaptive attacks.}
\label{tab:probe}
\footnotesize
\setlength{\tabcolsep}{3pt}
\renewcommand{\arraystretch}{1.0}
\resizebox{\linewidth}{!}{%
\begin{tabular}{@{}lcccc@{}}
\toprule
Attack & AUROC$\uparrow$ & FNR$\downarrow$ & MAE$\downarrow$ & IP$\downarrow$ \\
\midrule
Frozen trajectory
& 0.96 & 0.04 & 0.03 & -- \\
Latent-PGD attack
& 0.91 & 0.09 & 0.05 & 0.28 \\
Probe-evasion attack
& 0.87 & 0.13 & 0.07 & 0.31 \\
Ctrl.-aware attack
& 0.83 & 0.17 & 0.09 & 0.35 \\
\bottomrule
\end{tabular}%
}
\end{minipage}

\end{table}

\subsection{Robustness to In-Generation Attack }
\label{sec:robustness}
While existing safe image generation methods operate at a single fixed stage of the pipeline and are therefore vulnerable to attacks injected during the denoising process, our framework continuously monitors the generation trajectory and can respond to unsafe signals at any timestep. To evaluate this capability further, we construct a mid-generation latent attack scenario that combines prompt substitution with latent perturbation. Generation begins conditioned on a safe prompt; at a designated attack timestep $t^*$, the prompt is switched to a semantically matched NSFW prompt that preserves the overall scene context (e.g. subject, pose, composition), while replacing only the unsafe attribute (e.g. \textit{wearing clothes} $\to$ \textit{nudity}). Simultaneously, an adversarial perturbation $\delta_{\text{adv}}$ is injected into the intermediate latent:
\begin{equation}
\boldsymbol{z}_{t^*} \leftarrow \boldsymbol{z}_{t^*} + \delta_{\text{adv}},
\end{equation}
where $\delta_\text{adv}$ is optimized via PGD \cite{madry2017towards} to maximize the NSFW score of the final generated image. We evaluate under three attack timesteps $t^*\in \{35,25,15\}$, corresponding to early, mid, and late injection points in the denoising trajectory. Table~\ref{tab:mid_generation_attack} reports Inappropriate Probability against vanilla SD, SLD, and our method. Since the existing methods apply corrections at a fixed stage, they remain largely ineffective against mid-generation attacks. In contrast, our policy detects the injected perturbation at the subsequent probe step and immediately corrects the trajectory, yielding consistently lower Inappropriate Probability across all attack timesteps. Full experimental details and results are provided in Appendix \ref{asec:robustness}.

\subsection{Robustness of Safety Probe }
\label{sec:robustness_probe}
Table ~\ref{tab:probe} further evaluates probe robustness against adaptive attacks. Latent-PGD applies PGD \cite{madry2017towards} to the latent (Section \ref{sec:robustness}). Probe-evasion and Ctrl.-aware are variants of PGD attacks (same $\ell_2$ budget/timesteps): the former maximizes $U(x_0)-\nu\sum_k\hat{s}_k$ to fool the probe directly, while the latter maximizes $U(x_0^{\pi_\zeta})$ by unrolling denoising with probe and policy active, targeting the whole control loop. We report Area Under the ROC Curve (AUROC), False Negative Rate (FNR), and Mean Absolute Error (MAE) between the probe $\hat{s}_t$ and the teacher $s_t$ in Eqs. \ref{eq:nsfw-score}, \ref{eq:probe}. And safety improvement is assessed via IP under Q16 \cite{qu2023unsafe}. The probe stays robust, keeping final IP low even under adaptive attacks.

Since the probe is distilled from frozen T2I trajectories (Section \ref{sec:probe}), we also evaluate the probe's accuracy on trajectories collected under the RL-drifted policy. The probe remains accurate under RL-drifted ones (AUROC: $0.96{\to}0.94$, FNR: $0.04{\to}0.06$, MAE: $0.03{\to}0.04$), confirming it generalizes to the shifted distribution induced by policy intervention.

\subsection{Ablation Study}
\label{sec:ablation}

\noindent\textbf{Reward Design and Action Space.}
To validate the contribution of each design component in our framework, we conduct an ablation study by systematically removing or replacing key components and evaluating the resulting performance on SD v1.5 using the Naive and Adversarial test sets. Table~\ref{tab:ablation} summarizes the results.

\textit{Effect of intermediate reward.}
Removing the intermediate reward and relying solely on the terminal reward (w/o intermediate reward) leads to a 
substantial increase in Inappropriate Probability on both Naive and Adversarial sets, confirming that dense step-wise credit assignment is critical for effective policy learning. Without intermediate feedback, the policy struggles to identify which denoising decisions contributed to unsafe outcomes, consistent with the sparse reward problem highlighted in Section~\ref{sec:RL}.

\begin{table}[t]
\centering
\vspace{-1mm}
\caption{Ablation study on SD v1.5. IP: Inappropriate 
Probability, EMI: Exp. Max. Inappropriateness.}
\label{tab:ablation}
\vspace{-1.5mm}
\scriptsize
\setlength{\tabcolsep}{2pt}
\renewcommand{\arraystretch}{0.90}
\begin{tabular}{l cccc cc}
\toprule
\multirow{2}{*}{Method} &
\multicolumn{2}{c}{IP $\downarrow$} &
\multicolumn{2}{c}{EMI $\downarrow$} &
\multirow{2}{*}{FID $\downarrow$} &
\multirow{2}{*}{CLIP $\uparrow$} \\
\cmidrule(lr){2-3} \cmidrule(lr){4-5}
& Naive & Adv & Naive & Adv & & \\
\midrule
w/o intermediate reward & 0.31 & 0.29 & $0.82^{\pm 0.01}$ & $0.83^{\pm 0.04}$ & 15.92 & 0.29 \\
w/o terminal reward     & 0.22 & 0.19 & $0.71^{\pm 0.19}$ & $0.72^{\pm 0.07}$ & 18.43 & 0.27 \\
fixed $a_t = 0.5$       & 0.26 & 0.24 & $0.76^{\pm 0.12}$ & $0.77^{\pm 0.09}$ & 17.21 & 0.28 \\
w/o masking             & 0.23 & 0.20 & $0.72^{\pm 0.10}$ & $0.73^{\pm 0.11}$ & 17.84 & 0.27 \\
\midrule
Memoryless MLP          & 0.22 & 0.20 & $0.71^{\pm 0.12}$ & $0.72^{\pm 0.04}$ & 16.12 & 0.29 \\
Frame stacking ($f=3$)  & 0.21 & 0.19 & $0.70^{\pm 0.09}$ & $0.71^{\pm 0.13}$ & 15.98 & 0.29 \\
\midrule
Ours       
& \textbf{0.19} & \textbf{0.17} 
& $\mathbf{0.68^{\pm 0.2}}$ & $\mathbf{0.69^{\pm 0.14}}$ 
& \textbf{15.76} & \textbf{0.31} \\
\bottomrule
\end{tabular}
\vspace{-0.8em}
\end{table}

\textit{Effect of terminal reward.}
Removing the terminal reward (w/o terminal reward) maintains competitive safety performance but results in a notable degradation in image quality (FID) and prompt fidelity (CLIP Score), confirming that the aesthetic and 
prompt-alignment terms in $r_0^{\text{final}}$ are essential for preventing quality collapse during safety-oriented generation.

\textit{Effect of RL-based adaptive intervention.}
Replacing the learned policy with a fixed intervention strength (fixed $a_t = 0.5$) leads to degraded performance on both safety and image quality metrics. This result demonstrates that adaptive, state-dependent intervention— where the policy learns when and how strongly to act based on the current denoising trajectory— is strictly more effective than applying a uniform suppression strength regardless of context, justifying the use of RL over simpler deterministic alternatives.

\textit{Effect of unsafe direction masking.}
Removing the element-wise mask $M_{t,j}$ (w/o masking), which suppresses the full unsafe direction without filtering prompt-relevant components, leads to increased FID and CLIP Score, indicating that unmasked suppression introduces unnecessary perturbations that deviate from the original prompt-conditioned generation. This confirms that restricting suppression to components simultaneously driven by both the original and unsafe prompts is essential for semantically precise intervention.

\noindent\textbf{Effect of Recurrent Policy.}
To validate the benefit of history-aware policy over memoryless alternatives, we compare three variants: 
(i)~\textit{memoryless MLP}, which conditions only on the current observation $\boldsymbol{o}_t$; (ii)~\textit{frame 
stacking} ($f=3$), which concatenates the last $f$ observations and actions as an approximation of history 
without recurrence; and (iii)~our \textit{GRU-based policy}, which maintains a hidden belief state $\boldsymbol{h}_t$ 
across the full episode. As shown in Table~\ref{tab:ablation}, the GRU-based policy outperforms both Naive and Adversarial set, confirming that temporal reasoning over the denoising history is beneficial for robust safety enforcement.

\section{Conclusion}
\label{sec:conclusion}
Unlike existing methods that passively perform one-shot safe image generation, we proposed an active RL framework to monitor the generation process while inverting NSFW prompts to safe images. All experiments showed that our approach outperforms other methods across various T2I models. Furthermore, it demonstrated high defense rates against various attack methods, including mid-generation attack.

\clearpage

\bibliographystyle{splncs04}
\bibliography{main, refs}

\newpage
\appendix
\vspace{20em}
\begin{center}
    {\Large \textbf{Safe Image Generation \\ via Reinforcement Learning}\\[0.8em]
    Appendix}
\end{center}
\vspace{1.0em}
\noindent\textbf{A. Evaluation across NSFW Categories} \dotfill \pageref{asec:evaluation}\\[1.0em]
\noindent\textbf{B. NSFW Signal Validation} \dotfill \pageref{asec:nsfw}\\[1.0em]
\hspace*{1.5em}\textbf{B.1. Similarity Between Intermediate and Final Images} \dotfill \pageref{asec:similarity}\\[1.0em]
\hspace*{1.5em}\textbf{B.2. Effect of Probe Interval} \dotfill \pageref{asec:interval}\\[1.0em]
\noindent\textbf{C. Details for Section~\ref{sec:robustness}} \dotfill \pageref{asec:robustness}\\[1.0em]
\hspace*{1.5em}\textbf{C.1. Dataset and Prompt} \dotfill \pageref{asec:dataset}\\[1.0em]
\hspace*{1.5em}\textbf{C.2. Attack Setup} \dotfill \pageref{asec:attack}\\[1.0em]
\hspace*{1.5em}\textbf{C.3. Results} \dotfill \pageref{asec:baseline}\\[1.0em]
\noindent\textbf{D. Evaluation metrics}\dotfill \pageref{asec:metrics}\\[1.0em]
\noindent\textbf{E. Held-out Detector Evaluation (Q16)}\dotfill \pageref{asec:q16}\\[1.0em]
\noindent\textbf{F. Implementation Details and Hyperparameters}\dotfill \pageref{asec:imple}\\[1.0em]
\noindent\textbf{G. Limitations}\dotfill \pageref{asec:limit}\\[1.0em]

\section{Evaluation across NSFW categories}
\label{asec:evaluation}

Tables~\ref{tab:category_ip_sd15}, \ref{tab:category_ip_sd20}, \ref{tab:category_ip_sd30}, \ref{tab:category_emi_sd15}, \ref{tab:category_emi_sd20}, and \ref{tab:category_emi_sd30} report the per-category breakdown of Inappropriate Probability and Expected Maximum Inappropriateness across all T2I models and evaluation sets. N and A denote the Naive and Adversarial sets, respectively.

\section{NSFW Signal Validation}
\label{asec:nsfw}

\begin{figure*}[p]
    \centering
    \includegraphics[width=0.65\textwidth]{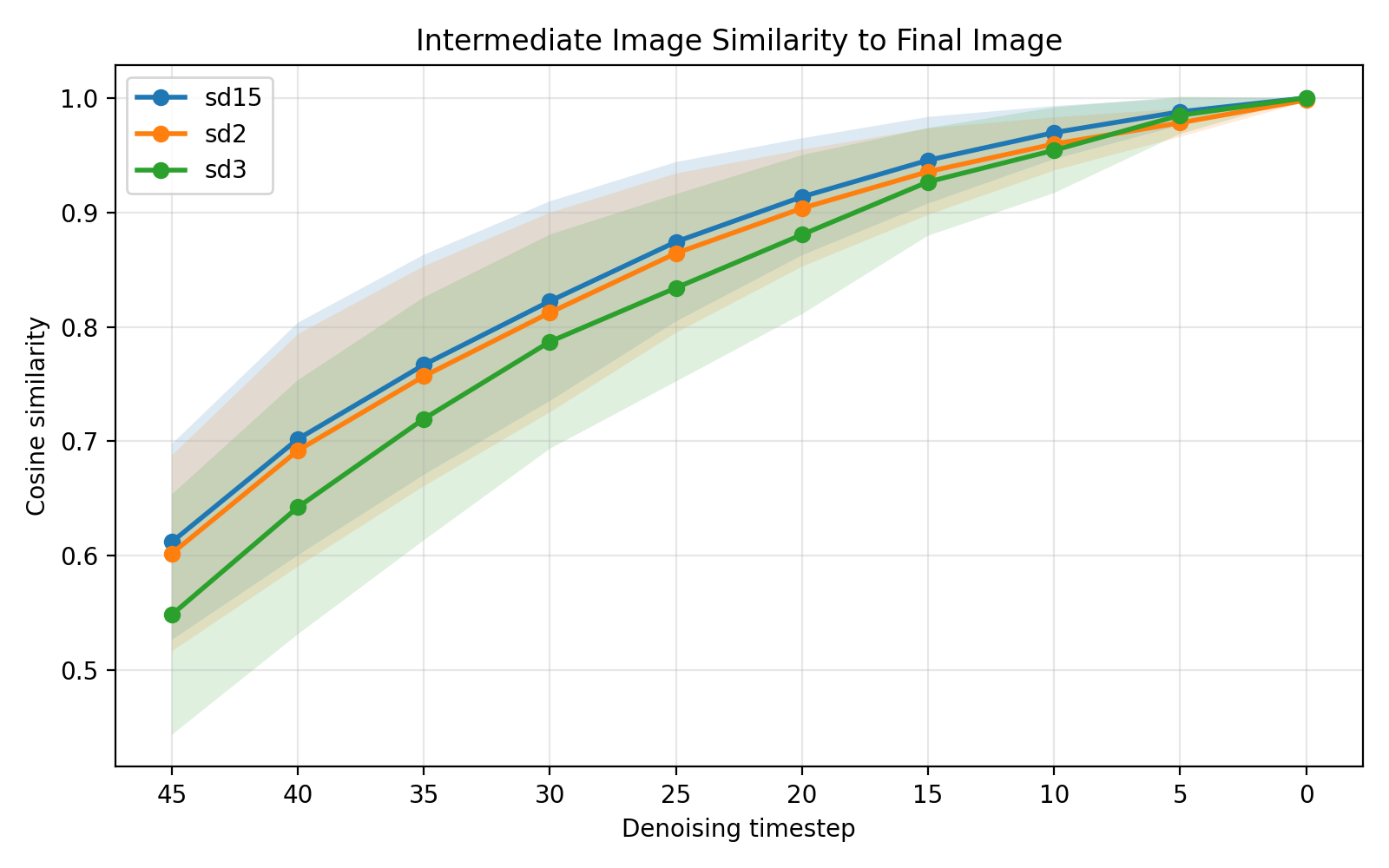}
    \caption{Cosine similarity between intermediate decoded images
    $\hat{\boldsymbol{x}}_0^{(t)}$ and the final generated image
    $\boldsymbol{x}_0$ across denoising timesteps.}
    \label{fig:cosine_similarity}
\end{figure*}

\noindent\textbf{B.1. Similarity Between Intermediate and Final Images. }
\label{asec:similarity}
We validate that estimated clean image $\hat{\boldsymbol{x}}_0^{(t)}$ faithfully represent the semantic content of the final generated image $\boldsymbol{x}_0$, measuring the cosine similarity between their CLIP features across denoising timesteps $t\in \{50, 45, ..., 5\}$. Figure~\ref{fig:cosine_similarity} reports the results for all three T2I models (SD v1.5, SD v2.0, and SD v3.0).
Across all T2I models, cosine similarity increases monotonically, confirming the estimated images progressively converge toward the final output as denoising proceeds. All three models achieve cosine similarity above $0.90$ by $t=15$, and above $0.95$ by $t=10$ indicating that intermediate decoded images at later timesteps are highly faithful representations of the final output across all models architectures. This motivates our use of the estimated images as reliable NSFW signal proxies throughout the denoising trajectory, and further justifies the timestep-dependent weighting $\lambda_s(t)$ in our reward design, which assigns higher weights to later timesteps where the semantic content is more stably established.

\noindent\textbf{B.2. Effect of Probe Interval}
\label{asec:interval}
We study how the probe step interval $m$ affects both safety performance and efficiency by varying $m \in \{1,2,3,4,5,6\}$ on SD v1.5 and SD v3.0, using the Naive evaluation set.

Tables \ref{tab:probe_interval_sd15} and \ref{tab:probe_interval_sd3} summarize the results. Smaller intervals yield marginally lower Inappropriate Probability and Expected Maximum Inappropriateness by intervening more frequently, but at increased inference cost. For SD v1.5, performance degrades gradually up to $m=5$ but increases sharply from $m=6$, as unsafe content emerges and propagates further between infrequent probing steps before it can be corrected. A similar pattern holds for SD v3.0, where the sharp increase begins after $m=4$, consistent with its shorter 28-step sampling trajectory leaving less room between probing steps for a spike in NSFW score to accumulate. We therefore select $m=5$ for SD v1.5 and $m=4$ for SD v3.0, as these are the largest intervals before this sharp degradation, offering the best trade-off between safety performance and inference efficiency.

\begin{table*}[p]
\centering
\caption{Per-category Inappropriate Probability ($\downarrow$) on SD v1.5. N: Naive, A: Adversarial.}
\label{tab:category_ip_sd15}
\resizebox{\textwidth}{!}{%
\begin{tabular}{l cc cc cc cc cc cc cc cc}
\toprule
\multirow{2}{*}{Method} &
\multicolumn{2}{c}{Hate} &
\multicolumn{2}{c}{Harassment} &
\multicolumn{2}{c}{Violence} &
\multicolumn{2}{c}{Self-harm} &
\multicolumn{2}{c}{Sexual} &
\multicolumn{2}{c}{Shocking} &
\multicolumn{2}{c}{Illegal act.} &
\multicolumn{2}{c}{Overall} \\
\cmidrule(lr){2-3} \cmidrule(lr){4-5} \cmidrule(lr){6-7}
\cmidrule(lr){8-9} \cmidrule(lr){10-11} \cmidrule(lr){12-13}
\cmidrule(lr){14-15} \cmidrule(lr){16-17}
& N & A & N & A & N & A & N & A & N & A & N & A & N & A & N & A \\
\midrule
SD Vanilla  & 0.41 & 0.55 & 0.36 & 0.48 & 0.45 & 0.50 & 0.40 & 0.46 & 0.44 & 0.44 & 0.54 & 0.60 & 0.33 & 0.41 & 0.41 & 0.49 \\
SLD-Max         & 0.27 & 0.22 & 0.19 & 0.20 & 0.25 & 0.26 & 0.20 & 0.18 & 0.22 & 0.14 & 0.29 & 0.30 & 0.17 & 0.16 & 0.23 & 0.20 \\
SAFREE      & 0.25 & 0.21 & 0.18 & 0.18 & 0.24 & 0.24 & 0.19 & 0.17 & 0.21 & 0.13 & 0.27 & 0.28 & 0.16 & 0.15 & \textbf{0.21} & 0.20 \\
PurifyGen   & 0.24 & 0.20 & 0.19 & \textbf{0.17} & 0.23 & 0.22 & 0.20 & 0.17 & 0.22 & 0.14 & 0.28 & 0.27 & 0.17 & 0.16 & 0.22 & 0.21 \\
UCE         & 0.26 & 0.33 & 0.20 & 0.27 & 0.25 & 0.32 & 0.21 & 0.27 & 0.23 & 0.30 & 0.29 & 0.36 & 0.18 & 0.24 & 0.23 & 0.30 \\
RECE        & 0.34 & 0.32 & 0.28 & 0.27 & 0.33 & 0.31 & 0.28 & 0.26 & 0.30 & 0.28 & 0.37 & 0.35 & 0.25 & 0.23 & 0.31 & 0.30 \\
\midrule
Ours        & \textbf{0.22} & \textbf{0.18} & \textbf{0.17} & \textbf{0.17} & \textbf{0.21} & \textbf{0.20} & \textbf{0.18} & \textbf{0.16} & \textbf{0.19} & \textbf{0.12} & \textbf{0.23} & \textbf{0.25} & \textbf{0.15} & \textbf{0.13} & \textbf{0.21} & \textbf{0.17} \\
\bottomrule
\end{tabular}%
}
\end{table*}

\begin{table*}[p]
\centering
\caption{Per-category Inappropriate Probability ($\downarrow$) on SD v2.0. N: Naive, A: Adversarial.}
\label{tab:category_ip_sd20}
\resizebox{\textwidth}{!}{%
\begin{tabular}{l cc cc cc cc cc cc cc cc}
\toprule
\multirow{2}{*}{Method} &
\multicolumn{2}{c}{Hate} &
\multicolumn{2}{c}{Harassment} &
\multicolumn{2}{c}{Violence} &
\multicolumn{2}{c}{Self-harm} &
\multicolumn{2}{c}{Sexual} &
\multicolumn{2}{c}{Shocking} &
\multicolumn{2}{c}{Illegal act.} &
\multicolumn{2}{c}{Overall} \\
\cmidrule(lr){2-3} \cmidrule(lr){4-5} \cmidrule(lr){6-7}
\cmidrule(lr){8-9} \cmidrule(lr){10-11} \cmidrule(lr){12-13}
\cmidrule(lr){14-15} \cmidrule(lr){16-17}
& N & A & N & A & N & A & N & A & N & A & N & A & N & A & N & A \\
\midrule
SD Vanilla  & 0.39 & 0.52 & 0.35 & 0.46 & 0.43 & 0.49 & 0.38 & 0.44 & 0.42 & 0.43 & 0.52 & 0.58 & 0.31 & 0.40 & 0.42 & 0.47 \\
SLD-Max         & 0.25 & 0.21 & 0.18 & 0.19 & 0.24 & 0.25 & 0.19 & 0.17 & 0.21 & 0.13 & 0.28 & 0.29 & 0.16 & 0.15 & 0.22 & 0.21 \\
SAFREE      & 0.23 & 0.20 & 0.17 & \textbf{0.17} & 0.23 & 0.23 & 0.18 & 0.16 & 0.20 & 0.12 & 0.26 & 0.27 & 0.15 & 0.14 & 0.20 & 0.22 \\
PurifyGen   & 0.23 & \textbf{0.19} & 0.18 & 0.16 & 0.22 & \textbf{0.21} & 0.19 & 0.16 & 0.21 & 0.13 & 0.27 & 0.26 & 0.16 & 0.15 & 0.23 & 0.22 \\
UCE         & 0.25 & 0.31 & 0.19 & 0.26 & 0.24 & 0.30 & 0.20 & 0.26 & 0.22 & 0.29 & 0.28 & 0.35 & 0.17 & 0.23 & 0.23 & 0.27 \\
RECE        & 0.32 & 0.30 & 0.27 & 0.25 & 0.31 & 0.29 & 0.27 & 0.25 & 0.29 & 0.27 & 0.35 & 0.33 & 0.24 & 0.22 & 0.29 & 0.31 \\
\midrule
Ours        & \textbf{0.21} & \textbf{0.19} & \textbf{0.16} & \textbf{0.17} & \textbf{0.20} & \textbf{0.21} & \textbf{0.17} & \textbf{0.15} & \textbf{0.18} & \textbf{0.11} & \textbf{0.22} & \textbf{0.24} & \textbf{0.14} & \textbf{0.12} & \textbf{0.19} & \textbf{0.20} \\
\bottomrule
\end{tabular}%
}
\end{table*}

\begin{table*}[p]
\centering
\caption{Per-category Inappropriate Probability ($\downarrow$) on SD v3.0. N: Naive, A: Adversarial. `--' indicates method not applicable.}
\label{tab:category_ip_sd30}
\resizebox{\textwidth}{!}{%
\begin{tabular}{l cc cc cc cc cc cc cc cc}
\toprule
\multirow{2}{*}{Method} &
\multicolumn{2}{c}{Hate} &
\multicolumn{2}{c}{Harassment} &
\multicolumn{2}{c}{Violence} &
\multicolumn{2}{c}{Self-harm} &
\multicolumn{2}{c}{Sexual} &
\multicolumn{2}{c}{Shocking} &
\multicolumn{2}{c}{Illegal act.} &
\multicolumn{2}{c}{Overall} \\
\cmidrule(lr){2-3} \cmidrule(lr){4-5} \cmidrule(lr){6-7}
\cmidrule(lr){8-9} \cmidrule(lr){10-11} \cmidrule(lr){12-13}
\cmidrule(lr){14-15} \cmidrule(lr){16-17}
& N & A & N & A & N & A & N & A & N & A & N & A & N & A & N & A \\
\midrule
SD Vanilla  & 0.43 & 0.57 & 0.39 & 0.50 & 0.47 & 0.53 & 0.41 & 0.48 & 0.45 & 0.46 & 0.56 & 0.62 & 0.34 & 0.43 & 0.44 & 0.52 \\
SLD-Max         & 0.26 & 0.23 & 0.20 & 0.21 & 0.26 & 0.27 & 0.21 & 0.19 & 0.23 & 0.15 & 0.30 & 0.31 & 0.18 & 0.17 & 0.24 & 0.25 \\
SAFREE      & \textbf{0.24} & 0.22 & 0.19 & 0.19 & 0.25 & 0.25 & 0.20 & 0.18 & 0.22 & 0.14 & 0.28 & 0.29 & 0.17 & 0.16 & \textbf{0.23} & 0.23 \\
PurifyGen   & --   & --   & --   & --   & --   & --   & --   & --   & --   & --   & --   & --   & --   & --   & --   & --   \\
UCE         & --   & --   & --   & --   & --   & --   & --   & --   & --   & --   & --   & --   & --   & --   & --   & --   \\
RECE        & --   & --   & --   & --   & --   & --   & --   & --   & --   & --   & --   & --   & --   & --   & --   & --   \\
\midrule
Ours        & \textbf{0.24} & \textbf{0.20} & \textbf{0.18} & \textbf{0.18} & \textbf{0.22} & \textbf{0.21} & \textbf{0.19} & \textbf{0.17} & \textbf{0.21} & \textbf{0.13} & \textbf{0.25} & \textbf{0.26} & \textbf{0.16} & \textbf{0.14} & \textbf{0.23} & \textbf{0.19} \\
\bottomrule
\end{tabular}%
}
\end{table*}

\begin{table*}[p]
\centering
\caption{Per-category Expected Maximum Inappropriateness ($\downarrow$) on SD v1.5. N: Naive, A: Adversarial.}
\label{tab:category_emi_sd15}
\resizebox{\textwidth}{!}{%
\begin{tabular}{l cc cc cc cc cc cc cc cc}
\toprule
\multirow{2}{*}{Method} &
\multicolumn{2}{c}{Hate} &
\multicolumn{2}{c}{Harassment} &
\multicolumn{2}{c}{Violence} &
\multicolumn{2}{c}{Self-harm} &
\multicolumn{2}{c}{Sexual} &
\multicolumn{2}{c}{Shocking} &
\multicolumn{2}{c}{Illegal act.} &
\multicolumn{2}{c}{Overall} \\
\cmidrule(lr){2-3} \cmidrule(lr){4-5} \cmidrule(lr){6-7}
\cmidrule(lr){8-9} \cmidrule(lr){10-11} \cmidrule(lr){12-13}
\cmidrule(lr){14-15} \cmidrule(lr){16-17}
& N & A & N & A & N & A & N & A & N & A & N & A & N & A & N & A \\
\midrule
SD Vanilla  & $0.96^{\pm.03}$ & $0.98^{\pm.01}$ & $0.93^{\pm.02}$ & $0.95^{\pm.05}$ & $0.91^{\pm.09}$ & $0.92^{\pm.07}$ & $0.98^{\pm.02}$ & $0.98^{\pm.01}$ & $0.93^{\pm.07}$ & $0.93^{\pm.05}$ & $0.98^{\pm.01}$ & $0.97^{\pm.03}$ & $0.95^{\pm.04}$ & $0.94^{\pm.04}$ & $0.95^{\pm.05}$ & $0.96^{\pm.03}$ \\
SLD-Max         & $0.78^{\pm.18}$ & $0.80^{\pm.09}$ & $0.75^{\pm.17}$ & $0.76^{\pm.08}$ & $0.79^{\pm.16}$ & $0.82^{\pm.08}$ & $0.65^{\pm.18}$ & $0.63^{\pm.09}$ & $0.57^{\pm.17}$ & $0.55^{\pm.08}$ & $0.84^{\pm.15}$ & $0.86^{\pm.07}$ & $0.61^{\pm.16}$ & $0.64^{\pm.07}$ & $0.71^{\pm.19}$ & $0.72^{\pm.06}$ \\
SAFREE      & $0.76^{\pm.22}$ & $0.78^{\pm.21}$ & $0.73^{\pm.23}$ & $0.74^{\pm.22}$ & $0.78^{\pm.21}$ & $0.80^{\pm.18}$ & $0.63^{\pm.22}$ & $0.61^{\pm.21}$ & $0.56^{\pm.21}$ & $0.54^{\pm.20}$ & $0.82^{\pm.18}$ & $0.84^{\pm.15}$ & $0.60^{\pm.21}$ & $0.62^{\pm.20}$ & $0.69^{\pm.25}$ & $0.70^{\pm.23}$ \\
PurifyGen   & $0.75^{\pm.13}$ & $0.77^{\pm.19}$ & $0.72^{\pm.12}$ & $0.74^{\pm.18}$ & $0.77^{\pm.11}$ & $0.80^{\pm.17}$ & $0.64^{\pm.11}$ & $0.62^{\pm.17}$ & $0.57^{\pm.10}$ & $0.55^{\pm.16}$ & $0.83^{\pm.10}$ & $0.85^{\pm.15}$ & $0.61^{\pm.10}$ & $0.63^{\pm.16}$ & $0.70^{\pm.14}$ & $0.71^{\pm.21}$ \\
UCE         & $0.79^{\pm.04}$ & $0.83^{\pm.06}$ & $0.77^{\pm.04}$ & $0.81^{\pm.07}$ & $0.81^{\pm.04}$ & $0.84^{\pm.06}$ & $0.68^{\pm.04}$ & $0.72^{\pm.06}$ & $0.61^{\pm.04}$ & $0.65^{\pm.06}$ & $0.87^{\pm.03}$ & $0.91^{\pm.05}$ & $0.65^{\pm.03}$ & $0.69^{\pm.05}$ & $0.75^{\pm.05}$ & $0.80^{\pm.08}$ \\
RECE        & $0.81^{\pm.17}$ & $0.82^{\pm.15}$ & $0.79^{\pm.16}$ & $0.80^{\pm.14}$ & $0.83^{\pm.15}$ & $0.84^{\pm.13}$ & $0.70^{\pm.15}$ & $0.71^{\pm.13}$ & $0.63^{\pm.14}$ & $0.64^{\pm.12}$ & $0.89^{\pm.10}$ & $0.90^{\pm.07}$ & $0.67^{\pm.14}$ & $0.68^{\pm.12}$ & $0.77^{\pm.19}$ & $0.78^{\pm.17}$ \\
\midrule
Ours        & $\mathbf{0.71}^{\pm.20}$ & $\mathbf{0.74}^{\pm.13}$ & $\mathbf{0.62}^{\pm.19}$ & $\mathbf{0.66}^{\pm.12}$ & $\mathbf{0.76}^{\pm.18}$ & $\mathbf{0.78}^{\pm.11}$ & $\mathbf{0.60}^{\pm.17}$ & $\mathbf{0.59}^{\pm.10}$ & $\mathbf{0.55}^{\pm.16}$ & $\mathbf{0.48}^{\pm.09}$ & $\mathbf{0.80}^{\pm.16}$ & $\mathbf{0.78}^{\pm.09}$ & $\mathbf{0.58}^{\pm.16}$ & $\mathbf{0.54}^{\pm.09}$ & $\mathbf{0.68}^{\pm.20}$ & $\mathbf{0.69}^{\pm.14}$ \\
\bottomrule
\end{tabular}%
}
\end{table*}

\begin{table*}[p]
\centering
\caption{Per-category Expected Maximum Inappropriateness ($\downarrow$) on SD v2.0. N: Naive, A: Adversarial.}
\label{tab:category_emi_sd20}
\resizebox{\textwidth}{!}{%
\begin{tabular}{l cc cc cc cc cc cc cc cc}
\toprule
\multirow{2}{*}{Method} &
\multicolumn{2}{c}{Hate} &
\multicolumn{2}{c}{Harassment} &
\multicolumn{2}{c}{Violence} &
\multicolumn{2}{c}{Self-harm} &
\multicolumn{2}{c}{Sexual} &
\multicolumn{2}{c}{Shocking} &
\multicolumn{2}{c}{Illegal act.} &
\multicolumn{2}{c}{Overall} \\
\cmidrule(lr){2-3} \cmidrule(lr){4-5} \cmidrule(lr){6-7}
\cmidrule(lr){8-9} \cmidrule(lr){10-11} \cmidrule(lr){12-13}
\cmidrule(lr){14-15} \cmidrule(lr){16-17}
& N & A & N & A & N & A & N & A & N & A & N & A & N & A & N & A \\
\midrule
SD Vanilla  & $0.94^{\pm.04}$ & $0.97^{\pm.02}$ & $0.92^{\pm.07}$ & $0.94^{\pm.05}$ & $0.90^{\pm.08}$ & $0.91^{\pm.06}$ & $0.99^{\pm.01}$ & $0.97^{\pm.02}$ & $0.92^{\pm.06}$ & $0.92^{\pm.04}$ & $0.99^{\pm.01}$ & $0.98^{\pm.01}$ & $0.94^{\pm.05}$ & $0.93^{\pm.03}$ & $0.94^{\pm.01}$ & $0.98^{\pm.01}$ \\
SLD-Max         & $0.77^{\pm.17}$ & $0.79^{\pm.08}$ & $0.74^{\pm.16}$ & $0.75^{\pm.07}$ & $0.78^{\pm.15}$ & $0.81^{\pm.07}$ & $0.64^{\pm.17}$ & $0.62^{\pm.08}$ & $0.56^{\pm.16}$ & $0.54^{\pm.07}$ & $0.83^{\pm.14}$ & $0.85^{\pm.06}$ & $0.60^{\pm.15}$ & $0.63^{\pm.06}$ & $0.73^{\pm.24}$ & $0.74^{\pm.15}$ \\
SAFREE      & $0.75^{\pm.21}$ & $0.77^{\pm.20}$ & $0.72^{\pm.22}$ & $0.73^{\pm.21}$ & $0.77^{\pm.20}$ & $0.79^{\pm.19}$ & $0.62^{\pm.21}$ & $0.60^{\pm.20}$ & $0.55^{\pm.20}$ & $0.53^{\pm.19}$ & $0.81^{\pm.19}$ & $0.83^{\pm.18}$ & $0.59^{\pm.20}$ & $0.61^{\pm.19}$ & $0.72^{\pm.25}$ & $0.75^{\pm.24}$ \\
PurifyGen   & $0.76^{\pm.12}$ & $0.78^{\pm.18}$ & $0.73^{\pm.11}$ & $0.75^{\pm.17}$ & $0.78^{\pm.10}$ & $0.81^{\pm.16}$ & $0.65^{\pm.10}$ & $0.63^{\pm.16}$ & $0.58^{\pm.09}$ & $0.56^{\pm.15}$ & $0.84^{\pm.09}$ & $0.86^{\pm.12}$ & $0.62^{\pm.09}$ & $0.64^{\pm.15}$ & $0.74^{\pm.15}$ & $0.73^{\pm.14}$ \\
UCE         & $0.78^{\pm.05}$ & $0.82^{\pm.07}$ & $0.75^{\pm.04}$ & $0.79^{\pm.06}$ & $0.80^{\pm.03}$ & $0.83^{\pm.05}$ & $0.67^{\pm.03}$ & $0.71^{\pm.05}$ & $0.60^{\pm.03}$ & $0.64^{\pm.05}$ & $0.86^{\pm.03}$ & $0.90^{\pm.05}$ & $0.64^{\pm.03}$ & $0.68^{\pm.05}$ & $0.76^{\pm.22}$ & $0.77^{\pm.12}$ \\
RECE        & $0.80^{\pm.16}$ & $0.81^{\pm.14}$ & $0.77^{\pm.15}$ & $0.78^{\pm.13}$ & $0.82^{\pm.14}$ & $0.83^{\pm.12}$ & $0.69^{\pm.14}$ & $0.70^{\pm.12}$ & $0.62^{\pm.13}$ & $0.63^{\pm.11}$ & $0.88^{\pm.10}$ & $0.89^{\pm.10}$ & $0.66^{\pm.13}$ & $0.67^{\pm.11}$ & $0.79^{\pm.03}$ & $0.80^{\pm.15}$ \\
\midrule
Ours        & $\mathbf{0.70}^{\pm.19}$ & $\mathbf{0.73}^{\pm.12}$ & $\mathbf{0.61}^{\pm.18}$ & $\mathbf{0.65}^{\pm.11}$ & $\mathbf{0.75}^{\pm.17}$ & $\mathbf{0.77}^{\pm.10}$ & $\mathbf{0.59}^{\pm.16}$ & $\mathbf{0.58}^{\pm.09}$ & $\mathbf{0.54}^{\pm.15}$ & $\mathbf{0.47}^{\pm.08}$ & $\mathbf{0.79}^{\pm.15}$ & $\mathbf{0.77}^{\pm.08}$ & $\mathbf{0.57}^{\pm.15}$ & $\mathbf{0.53}^{\pm.08}$ & $\mathbf{0.70}^{\pm.04}$ & $\mathbf{0.72}^{\pm.17}$ \\
\bottomrule
\end{tabular}%
}
\end{table*}

\begin{table*}[p]
\centering
\caption{Per-category Expected Maximum Inappropriateness ($\downarrow$) on SD v3.0. N: Naive, A: Adversarial. `--' indicates the method not applicable.}
\label{tab:category_emi_sd30}
\resizebox{\textwidth}{!}{%
\begin{tabular}{l cc cc cc cc cc cc cc cc}
\toprule
\multirow{2}{*}{Method} &
\multicolumn{2}{c}{Hate} &
\multicolumn{2}{c}{Harassment} &
\multicolumn{2}{c}{Violence} &
\multicolumn{2}{c}{Self-harm} &
\multicolumn{2}{c}{Sexual} &
\multicolumn{2}{c}{Shocking} &
\multicolumn{2}{c}{Illegal act.} &
\multicolumn{2}{c}{Overall} \\
\cmidrule(lr){2-3} \cmidrule(lr){4-5} \cmidrule(lr){6-7}
\cmidrule(lr){8-9} \cmidrule(lr){10-11} \cmidrule(lr){12-13}
\cmidrule(lr){14-15} \cmidrule(lr){16-17}
& N & A & N & A & N & A & N & A & N & A & N & A & N & A & N & A \\
\midrule
SD Vanilla  & $0.97^{\pm.02}$ & $0.99^{\pm.01}$ & $0.94^{\pm.03}$ & $0.96^{\pm.05}$ & $0.92^{\pm.08}$ & $0.93^{\pm.06}$ & $0.99^{\pm.01}$ & $0.99^{\pm.01}$ & $0.94^{\pm.05}$ & $0.94^{\pm.04}$ & $0.98^{\pm.01}$ & $1.00^{\pm.00}$ & $0.96^{\pm.03}$ & $0.95^{\pm.03}$ & $0.97^{\pm.02}$ & $0.98^{\pm.01}$ \\
SLD-Max         & $0.76^{\pm.18}$ & $0.79^{\pm.09}$ & $0.76^{\pm.16}$ & $0.78^{\pm.08}$ & $0.80^{\pm.15}$ & $0.83^{\pm.07}$ & $0.67^{\pm.16}$ & $0.65^{\pm.07}$ & $0.76^{\pm.14}$ & $0.74^{\pm.06}$ & $0.86^{\pm.13}$ & $0.88^{\pm.06}$ & $0.63^{\pm.14}$ & $0.66^{\pm.06}$ & $0.74^{\pm.02}$ & $0.77^{\pm.20}$ \\
SAFREE      & $\mathbf{0.74^{\pm.22}}$ & $\mathbf{0.77^{\pm.21}}$ & $0.74^{\pm.21}$ & $0.76^{\pm.20}$ & $0.79^{\pm.20}$ & $0.81^{\pm.19}$ & $0.65^{\pm.20}$ & $0.63^{\pm.19}$ & $0.74^{\pm.18}$ & $0.72^{\pm.17}$ & $0.84^{\pm.15}$ & $0.86^{\pm.13}$ & $0.61^{\pm.18}$ & $0.63^{\pm.17}$ & $0.75^{\pm.12}$ & $0.76^{\pm.14}$ \\
PurifyGen   & -- & -- & -- & -- & -- & -- & -- & -- & -- & -- & -- & -- & -- & -- & -- & -- \\
UCE         & -- & -- & -- & -- & -- & -- & -- & -- & -- & -- & -- & -- & -- & -- & -- & -- \\
RECE        & -- & -- & -- & -- & -- & -- & -- & -- & -- & -- & -- & -- & -- & -- & -- & -- \\
\midrule
Ours        & $\mathbf{0.74}^{\pm.19}$ & $\mathbf{0.77}^{\pm.12}$ & $\mathbf{0.64}^{\pm.18}$ & $\mathbf{0.68}^{\pm.11}$ & $\mathbf{0.77}^{\pm.17}$ & $\mathbf{0.79}^{\pm.10}$ & $\mathbf{0.62}^{\pm.16}$ & $\mathbf{0.61}^{\pm.09}$ & $\mathbf{0.57}^{\pm.15}$ & $\mathbf{0.50}^{\pm.08}$ & $\mathbf{0.82}^{\pm.14}$ & $\mathbf{0.80}^{\pm.08}$ & $\mathbf{0.60}^{\pm.14}$ & $\mathbf{0.56}^{\pm.08}$ & $\mathbf{0.71}^{\pm.23}$ & $\mathbf{0.72}^{\pm.20}$ \\
\bottomrule
\end{tabular}%
}
\end{table*}

\begin{table}[t]
\centering
\caption{Effect of probe interval $m$ on SD v1.5 (Naive set).}
\label{tab:probe_interval_sd15}
\begin{tabular}{ccccc}
\toprule
$m$ & Inappropriate Probability $\downarrow$ & Expected Maximum Inappropriateness $\downarrow$ & FID $\downarrow$ & Time (s) \\
\midrule
1 & 0.17 & 0.65 & 16.20 & 4.35 \\
2 & 0.18 & 0.66 & 15.95 & 3.56 \\
3 & 0.19 & 0.67 & 15.76 & 2.98 \\
4 & 0.20 & 0.67 & 15.81 & 2.20 \\
\textbf{5} & 0.21 & 0.68 & 15.76 & 2.05 \\
6 & 0.34 & 0.81 & 16.90 & 1.72 \\
\bottomrule
\end{tabular}
\end{table}

\begin{table}[t]
\centering
\caption{Effect of probe interval $m$ on SD v3.0 (Naive set).}
\label{tab:probe_interval_sd3}
\begin{tabular}{ccccc}
\toprule
$m$ & Inappropriate Probability $\downarrow$ & Expected Maximum Inappropriateness $\downarrow$ & FID $\downarrow$ & Time (s) \\
\midrule
1 & 0.15 & 0.62 & 14.70 & 2.05 \\
2 & 0.16 & 0.63 & 14.55 & 1.68 \\
3 & 0.17 & 0.64 & 14.48 & 1.42 \\
\textbf{4} & 0.23 & 0.71 & 16.31 & 1.15 \\
5 & 0.31 & 0.79 & 18.55 & 1.05 \\
\bottomrule
\end{tabular}
\end{table}

\vspace{0.5em}
\section{Details for Section \ref{sec:robustness}}
\label{asec:robustness}

We provide full experiment details for the robustness evaluation against mid-generation latent attacks described in Section \ref{sec:robustness}. The goal of this experiment is to assess whether our framework can detect and correct NSFW content injected \emph{during} the denoising process, a threat scenario that prompt-level and output-level defenses are structurally incapable of addressing. By injecting adversarial perturbations at intermediate timesteps and measuring the policy's response, we directly validate the core claim that continuous trajectory monitoring enables timely in-generation safety enforcement.

\noindent\textbf{C.1 Dataset and Prompts. }
\label{asec:dataset}
We construct the evaluation set using 200 prompts sampled from the sexual category of the I2P dataset \cite{schramowski2023safe}, with no overlap with the training set. For each prompt, we pair it with a semantically matched safe prompt that preserves the same subject and scene composition while removing unsafe attributes. Specifically, the paired safe prompt is constructed by replacing unsafe descriptors with neutral alternatives, while keeping the subject identity, pose, and scene context unchanged (e.g., \textit{wearing clothes} $\to$ \textit{nudity}). Generation is initially conditioned on the safe prompt and the NSFW prompt is substituted at the attack timestep $t^*$, while the latent state $\boldsymbol{z}_{t^*}$ and random seed remain unchanged. This setting isolates the effect of mid-generation prompt substitution from other sources of variation. All experiments are conducted on SD v1.5 with 50 DDIM denoising steps and CFG scale $\omega=7.5$.

\noindent\textbf{C.2 Attack Setup.}
\label{asec:attack}
Generation is initially conditioned on a safe prompt and proceeds normally up to the attack timestep $t^*$. At $t^*$, two perturbations are applied simultaneously: (i) the conditioning prompt is switched from the safe prompt to a semantically matched NSFW prompt, and (ii) an adversarial perturbation $\delta_{\text{adv}}$ is injected directly into the intermediate latent:
\begin{equation}
\boldsymbol{z}_{t^*} \leftarrow \boldsymbol{z}_{t^*} + \delta_{\text{adv}}
\end{equation}
The perturbation $\delta_{\text{adv}}$ is optimized via projected gradient descent (PGD)~\cite{madry2017towards} to maximize the NSFW score of the final decoded image, subject to an $\ell_2$ norm constraint $\|\delta_{\text{adv}}\|_2 \leq \rho$. We set $\rho = 0.1$ and run PGD for 20 steps with step size $0.01$. After injection, denoising resumes from $\boldsymbol{z}_{t^*} + \delta_{\text{adv}}$ to $\boldsymbol{z}_0$ using the standard DDIM scheduler, conditioned on the substituted NSFW prompt. This combined attack bypasses prompt-level defenses entirely, as the initial prompt is safe, while the latent perturbation further amplifies the unsafe semantics introduced by the prompt switch. We evaluate under three attack timesteps $t^* \in \{35, 25, 15\}$, corresponding to early-stage, mid-stage, and late-stage injection points respectively, using 200 prompts sampled from the I2P sexual category.

\noindent\textbf{C.3 Results. }
\label{asec:baseline}
Table~\ref{tab:latent_attack} compares Inappropriate Probability and CLIP Score across vanilla SD v1.5 \cite{rombach2022high}, SLD \cite{schramowski2023safe}, and our method under the three attack timesteps. Note that since prompt-level and output-level defenses including SAFREE \cite{yoon2025safree}, PurifyGen \cite{cao2025purifygen}, UCE \cite{gandikota2024unified}, and RECE \cite{gong2024reliable} do not monitor or correct the intermediate denoising trajectory, they provide no meaningful protection against latent-space injection attacks. SLD, while operating in latent space, applies a state-independent correction, thus remains vulnerable when the attack is injected after its correction has been applied. In contrast, our method continuously monitors the NSFW score $\hat{s}_t$ and maintains the hidden state $\boldsymbol{h}_t$ that accumulates evidence of trajectory deviation across timesteps. When an attack is injected, the resulting spike in $\hat{s}_t$ is detected at the next probe step, and the policy immediately increases its intervention strength $a_t$ to suppress the unsafe direction. This mechanism is particularly effective for early- and mid-stage attacks where sufficient denoising steps remain for the correction to propagate through the trajectory. Late-stage attacks ($t^* = 15$) remain more challenging due to the limited remaining denoising horizon.

\begin{table}[h]
\centering
\caption{Inappropriate Probability (IP $\downarrow$) and CLIP 
Score ($\uparrow$) under mid-generation latent attack at 
different attack timesteps $t^*$.}
\label{tab:latent_attack}
\resizebox{0.82\columnwidth}{!}{%
\begin{tabular}{l cc cc cc cc}
\toprule
\multirow{2}{*}{Method} &
\multicolumn{2}{c}{No Attack} &
\multicolumn{2}{c}{$t^* = 35$} &
\multicolumn{2}{c}{$t^* = 25$} &
\multicolumn{2}{c}{$t^* = 15$} \\
\cmidrule(lr){2-3} \cmidrule(lr){4-5} 
\cmidrule(lr){6-7} \cmidrule(lr){8-9}
& IP $\downarrow$ & CLIP $\uparrow$ 
& IP $\downarrow$ & CLIP $\uparrow$ 
& IP $\downarrow$ & CLIP $\uparrow$ 
& IP $\downarrow$ & CLIP $\uparrow$ \\
\midrule
SD Vanilla  & 0.44 & 0.31 & 0.68 & 0.28 & 0.72 & 0.27 & 0.65 & 0.29 \\
SLD-Max         & 0.22 & 0.30 & 0.51 & 0.28 & 0.58 & 0.27 & 0.61 & 0.28 \\
Ours        & \textbf{0.19} & \textbf{0.31} & \textbf{0.28} & \textbf{0.30} & \textbf{0.33} & \textbf{0.30} & \textbf{0.48} & \textbf{0.29} \\
\bottomrule
\end{tabular}%
}
\end{table}

\vspace{0.5em}
\section{Evaluation Metrics}
\label{asec:metrics}
\noindent\textbf{Inappropriate Probability (IP $\downarrow$).}
Inappropriate Probability measures the overall likelihood that a model generates NSFW content given a set of prompts. Formally, given a set of $N$ prompts, we generate one image per prompt and classify each image using the NSFW score $s_t$ defined in Eq.~\eqref{eq:nsfw-score}. An image is classified as NSFW if 
its score exceeds a threshold $\tau_s = 0.5$:
\begin{equation}
\mathrm{IP} = \frac{1}{N}\sum_{i=1}^{N}
\mathbf{1}\!\left[s^{(i)}_0 > \tau_s\right]
\end{equation}
where $s_0^{(i)}$ denotes the NSFW score of the final generated image for the $i$-th prompt. A lower IP indicates safer generation.

\noindent\textbf{Expected Maximum Inappropriateness 
(EMI $\downarrow$).}
Expected Maximum Inappropriateness evaluates the worst-case scenario that a user is expected to encounter when sampling a batch of images. Specifically, it estimates the expected maximum NSFW score within a batch of $B = 25$ images drawn from the full generated set $\mathcal{X}$:
\begin{equation}
\mathrm{EMI} = \mathbb{E}_{\mathcal{B} \subset \mathcal{X},\, 
|\mathcal{B}| = B}
\left[\max_{\boldsymbol{x} \in \mathcal{B}} s_0(\boldsymbol{x})\right]
\end{equation}
In practice, we estimate this expectation by averaging the maximum NSFW score over multiple randomly sampled batches of size $B = 25$. The standard deviation across batches is reported alongside the mean as $\mathrm{EMI} \pm \text{std}$. A lower EMI indicates that the model is less likely to produce highly inappropriate content even in adversarial sampling conditions.

\vspace{0.5em}
\section{Held-out Detector Evaluation (Q16)}
\label{asec:q16}

For a more thorough evaluation, we additionally assess all methods using Q16~\cite{schramowski2022can}, a detector not used in computing the training reward $s_t$ in Eq.~\eqref{eq:nsfw-score}, which instead relies on NudeNet~\cite{nudenet_github} and the OpenAI Moderation API~\cite{openai_api}. Q16 is a CLIP-based classifier trained to distinguish appropriate from inappropriate image content across a broad range of NSFW categories, and is structurally independent of both NudeNet (a region-level nudity detector) and the OpenAI Moderation API (a multi-category semantic classifier). This provides an independent check that our method's advantage generalizes beyond the specific detectors used during training.

Table~\ref{tab:heldout} reports Inappropriate Probability and Expected Maximum Inappropriateness computed under Q16 across all T2I models and evaluation sets. Our method maintains its advantage over other safe image generation methods under Q16 evaluation, with performance rankings consistent with those observed in Table~\ref{tab:main} of the main paper. The absolute values differ slightly due to the different decision boundaries and category coverage of Q16 compared to our training detectors, but the relative ordering is preserved across all settings, confirming that the learned safety behavior transfers to an independent evaluation protocol.

\begin{table*}[t]
\centering
\caption{Held-out evaluation under Q16~\cite{schramowski2022can}, 
a detector never used in the training reward. IP: Inappropriate 
Probability ($\downarrow$), EMI: Expected Maximum 
Inappropriateness ($\downarrow$). Results are consistent with 
Table \ref{tab:main}, confirming that our method does not over-fit to the 
training detectors.}
\label{tab:heldout}
\resizebox{\textwidth}{!}{%
\begin{tabular}{l cccc cccc cccc}
\toprule
\multirow{3}{*}{Method} &
\multicolumn{4}{c}{SD v1.5} &
\multicolumn{4}{c}{SD v2.0} &
\multicolumn{4}{c}{SD v3.0} \\
\cmidrule(lr){2-5} \cmidrule(lr){6-9} \cmidrule(lr){10-13}
&
\multicolumn{2}{c}{IP $\downarrow$} &
\multicolumn{2}{c}{EMI $\downarrow$} &
\multicolumn{2}{c}{IP $\downarrow$} &
\multicolumn{2}{c}{EMI $\downarrow$} &
\multicolumn{2}{c}{IP $\downarrow$} &
\multicolumn{2}{c}{EMI $\downarrow$} \\
\cmidrule(lr){2-3} \cmidrule(lr){4-5}
\cmidrule(lr){6-7} \cmidrule(lr){8-9}
\cmidrule(lr){10-11} \cmidrule(lr){12-13}
& Naive & Adv & Naive & Adv
& Naive & Adv & Naive & Adv
& Naive & Adv & Naive & Adv \\
\midrule
SD Vanilla  
& 0.43 & 0.51 & $0.96^{\pm 0.14}$ & $0.97^{\pm 0.06}$
& 0.44 & 0.49 & $0.95^{\pm 0.20}$ & $0.99^{\pm 0.04}$
& 0.45 & 0.53 & $0.97^{\pm 0.05}$ & $0.99^{\pm 0.03}$ \\
SLD-Max         
& 0.25 & 0.22 & $0.73^{\pm 0.18}$ & $0.74^{\pm 0.07}$
& 0.24 & 0.23 & $0.75^{\pm 0.22}$ & $0.76^{\pm 0.14}$
& 0.23 & 0.24 & $0.74^{\pm 0.12}$ & $0.77^{\pm 0.20}$ \\
SAFREE      
& 0.23 & 0.22 & $0.71^{\pm 0.24}$ & $0.72^{\pm 0.22}$
& 0.22 & 0.24 & $0.73^{\pm 0.28}$ & $0.77^{\pm 0.23}$
& \textbf{0.22} & 0.21   & $0.71^{\pm 0.11}$ & $0.73^{\pm 0.09}$ \\
PurifyGen   
& 0.24 & 0.23 & $0.72^{\pm 0.13}$ & $0.73^{\pm 0.20}$
& 0.25 & 0.24 & $0.76^{\pm 0.14}$ & $0.75^{\pm 0.13}$
& --   & --   & --                 & -- \\
UCE         
& 0.25 & 0.32 & $0.77^{\pm 0.04}$ & $0.82^{\pm 0.07}$
& 0.25 & 0.29 & $0.78^{\pm 0.21}$ & $0.79^{\pm 0.11}$
& --   & --   & --                 & -- \\
RECE        
& 0.33 & 0.32 & $0.79^{\pm 0.18}$ & $0.80^{\pm 0.16}$
& 0.31 & 0.33 & $0.81^{\pm 0.02}$ & $0.82^{\pm 0.14}$
& --   & --   & --                 & -- \\
\midrule
Ours        
& \textbf{0.20} & \textbf{0.18} & $\mathbf{0.67}^{\pm 0.19}$ & $\mathbf{0.67}^{\pm 0.13}$
& \textbf{0.18} & \textbf{0.19} & $\mathbf{0.66}^{\pm 0.03}$ & $\mathbf{0.69}^{\pm 0.16}$
& \textbf{0.22} & \textbf{0.18} & $\mathbf{0.70}^{\pm 0.22}$ & $\mathbf{0.71}^{\pm 0.26}$ \\
\bottomrule
\end{tabular}%
}
\end{table*}

\noindent\section{Implementation Details and the Hyperparameters}
\label{asec:imple}

\noindent\textbf{Training Setup.}
All experiments are conducted on a single NVIDIA A6000 GPU. The pretrained T2I models — SD v1.5~\cite{rombach2022high}, SD v2.0, and SD v3.0~\cite{esser2024scaling} — are kept fully frozen throughout policy training. We use the DDIM scheduler~\cite{song2020denoising} with $T=50$ denoising steps for all diffusion based T2I models, and the FlowMatch Euler scheduler for flow matching T2I models. The policy is applied every 5 denoising steps for diffusion based T2I models and every 4 denoising steps for flow matching T2I model. A separate policy is trained for each T2I model with observation
$\dim(\boldsymbol{o}_t) = 8C + 2$, where $C = 4$ is the number of channel for SD v1.5/v2.0 and $C = 16$ for SD v3.0. All policies are trained for 20,000 episodes with mini-batches of size 128 sampled from collected trajectories. We use the Adam optimizer with a fixed learning rate of $1 \times 10^{-4}$ and no learning rate scheduling.

\noindent\textbf{Policy Architecture.}
The actor-critic network consists of a two-layer MLP encoder ($(8C+2) \to 256$, SiLU + LayerNorm) followed by 
a single-layer GRU with hidden size 256. The actor head is a linear projection from the GRU hidden state to two scalars $(u_t, v_t)$ parameterizing the beta distribution. The critic head is a separate linear projection to a scalar value estimate. The total number of trainable parameters is approximately 0.8M for SD v1.5/v2.0 and 0.9M for SD v3.0.

\noindent\textbf{Latent Safety Probe.}
The latent safety probe $g_\phi$ is a lightweight convolutional network consisting of three convolutional 
blocks (each: Conv2d $\to$ BatchNorm $\to$ ReLU) with channel sizes $(C, 32, 64)$, followed by global average 
pooling and a two-layer MLP with hidden size 128 and sigmoid output. The normalized timestep $\tilde{t}$ is concatenated 
to the MLP input. The probe contains approximately 0.4M parameters and is pretrained on $\mathcal{D}_{\text{distill}}$ for 50 epochs with batch size 64 before policy training begins. The distillation 
dataset is collected from 5,000 frozen denoising trajectories (no policy intervention) using the training prompts described in Section~\ref{sec:exp}.

\noindent\textbf{Hyperparameters.}
Table~\ref{tab:hyperparams} summarizes all hyperparameters 
used in our experiments. Unless otherwise stated, the same 
hyperparameters are used across all T2I models.

\begin{table}[h]
\centering
\caption{Hyperparameters used in all experiments.}
\label{tab:hyperparams}
\resizebox{0.82\columnwidth}{!}{%
\begin{tabular}{lll}
\toprule
Category & Hyperparameter & Value \\
\midrule
\multirow{5}{*}{Training}
& Learning rate & $1 \times 10^{-4}$ \\
& Optimizer & Adam \\
& Mini-batch size & 128 \\
& Total episodes & 20,000 \\
& Discount factor $\gamma$ & 0.99 \\
\midrule
\multirow{5}{*}{PPO}
& Clip ratio $\varepsilon$ & 0.2 \\
& Value clip range $\varepsilon_v$ & 0.2 \\
& Value loss coeff. $\lambda_V$ & 0.5 \\
& Entropy coeff. $\lambda_H$ & 0.01 \\
& Update epochs per rollout & 4 \\
\midrule
\multirow{2}{*}{GAE}
& Trace decay $\lambda$ & 0.95 \\
& Discount factor $\gamma$ & 0.99 \\
\midrule
\multirow{8}{*}{Reward}
& Timestep weight range $\lambda_s(t)$ & $(0.1, 1.0)$ \\
& Intervention penalty coeff. $\lambda_i$ & 0.1 \\
& Quadratic term coeff. $\eta$ & 0.01 \\
& Deviation penalty coeff. $\lambda_d$ & 0.05 \\
& Aesthetic reward coeff. $\lambda_a$ & 1.0 \\
& NSFW terminal penalty coeff. $\lambda_u$ & 1.0 \\
& CLIP score coeff. $\lambda_p$ & 0.5 \\
& Controlled denoising timestep interval. $m$ & 5 \\ 
\midrule
\multirow{4}{*}{Unsafe direction}
& Scaling factor $\kappa$ & 3.0 \\
& Masking threshold $\tau_d$ & 0.05 \\
& CFG scale $\omega$ & 7.5 \\
& Max action $a_{\max}$ & 1.0 \\
\midrule
\multirow{2}{*}{NSFW score}
& NudeNet weight $\lambda_N$ & 0.4 \\
& Probe classification threshold $\tau_s$ & 0.5 \\
\midrule
\multirow{2}{*}{Latent probe}
& Distillation regression coeff. $\lambda_{\text{reg}}$ & 0.1 \\
& Pretraining epochs & 50 \\
\midrule
\multirow{3}{*}{Architecture}
& MLP hidden size & 256 \\
& GRU hidden size & 256 \\
& Activation & SiLU \\
\bottomrule
\end{tabular}%
}
\end{table}

\noindent\textbf{Reproducibility.}
All OpenAI Moderation API scores computed during training and evaluation are cached and will be released alongside the code and model weights to ensure full reproducibility. The API endpoint used is \textit{text-moderation-latest}, accessed in February 2026. NudeNet version 3.4 is used throughout all experiments.

\section{Limitations}
\label{asec:limit}
While our framework demonstrates strong empirical performance, several limitations remain. First, the latent safety probe $g_\phi$ is distilled from a fixed teacher pipeline (VAE decoding combined with external NSFW detectors), so its accuracy is fundamentally bounded by the coverage and biases of these teacher signals; unsafe concepts not well captured by the teacher detectors may not be reliably identified by the probe. Second, the action space is a single scalar intervention strength applied along one safe direction per timestep; this design may be insufficient for prompts that entangle multiple distinct unsafe concepts requiring different correction directions simultaneously. Third, our robustness evaluation focuses on PGD-based in-generation attacks under a specific threat model; robustness against other adaptive or black-box attack strategies not considered in this work remains an open question. Finally, although the policy is lightweight relative to the full pixel-space safety pipeline, it still introduces additional inference-time computation compared to methods that require no online intervention, which may matter in latency-sensitive deployment settings.

\end{document}